\documentclass[11pt]{article}

\usepackage[final]{acl}

\usepackage{times}
\usepackage{latexsym}

\usepackage[T1]{fontenc}

\usepackage[utf8]{inputenc}

\usepackage{microtype}

\usepackage{inconsolata}

\usepackage{graphicx}

\usepackage[utf8]{inputenc} 
\usepackage[T1]{fontenc}    
\usepackage{hyperref}       
\usepackage{url}            
\usepackage{booktabs}       
\usepackage{amsfonts}       
\usepackage{nicefrac}       
\usepackage{microtype}      
\usepackage{tabularx}
\usepackage{graphicx}
\usepackage{adjustbox}
\usepackage{multirow}
\usepackage{caption}
\usepackage{subcaption}
\usepackage[table,xcdraw]{xcolor}
\usepackage{enumitem}
\usepackage{wrapfig}
\usepackage{graphicx}
\usepackage{makecell}
\usepackage{multirow}
\usepackage{placeins}
\usepackage{float}
\usepackage{tcolorbox}
\tcbuselibrary{listingsutf8}
\usepackage{tocloft}
\usepackage{longtable} 
\usepackage{array}
\usepackage{fontawesome5}

\usepackage{flushend}

\title{Interpretable Traces, Unexpected Outcomes: Investigating the Disconnect in Trace-Based Knowledge Distillation}

\author{Siddhant Bhambri\thanks{Work done while a PhD student at ASU, currently at Samsung Research America}, Upasana Biswas, Subbarao Kambhampati \\
   School of Computing \& AI, Arizona State University \\
   \{siddhantbhambri, ubiswas2, rao\}@asu.edu
}

\begin{document}
\maketitle
\begin{abstract}
Recent advances in reasoning-oriented Large Language Models (LLMs) have been driven by the introduction of Chain-of-Thought (CoT) traces, where models generate intermediate reasoning traces before producing an answer. These traces, as in DeepSeek R1, are not only used to guide model inference but also serve as supervision signals for Knowledge Distillation (KD) to improve smaller models. A prevailing but under-examined implicit assumption is that these CoT traces when emitted at inference time are both semantically correct and interpretable for the end-users. While there are reasons to believe that these intermediate tokens help improve solution accuracy, in this work, we question their validity (semantic correctness) and interpretability to the end user. To isolate the effect of trace semantics, we design experiments in the Question Answering (QA) domain using a rule-based problem decomposition method. This enables us to create Supervised Fine-Tuning (SFT) datasets for LLMs where - each QA problem is paired with either verifiably correct or incorrect CoT traces, while always providing the correct final solution. Trace correctness at inference time is then evaluated by checking the accuracy of every sub-step in decomposed reasoning chains. To assess end-user interpretability, we finetune LLMs with three additional types of CoT traces: R1 traces, R1 trace summaries, and post-hoc explanations of R1 traces. We further conduct a human-subject study with 100 participants asking them to rate the interpretability of each trace type on a standardized Likert scale. Our experiments reveal two key findings - (1) CoT trace correctness is not reliably correlated with the model’s generation of correct final answers: correct traces led to correct solutions only for 28\% test-set problems while incorrect traces don't necessarily degrade solution accuracy. (2) In end-user interpretability studies, fine-tuning on verbose R1 traces produced the best model performance but these traces were rated as least interpretable by users, scoring on average 3.39 for interpretability and 4.59 for cognitive load metrics on a 5-point Likert scale. In contrast, the decomposed traces that are judged significantly more interpretable don't lead to comparable solution accuracy. Together, these findings challenge the assumption in question suggesting that researchers and practitioners should decouple model supervision objectives from end-user-facing trace design.\footnote{%
  \faGlobe\ \href{https://sbhambr1.github.io/trace-check-qa-paper/}{\texttt{Project Webpage}}
  \faGithub\ \href{https://github.com/sbhambr1/Trace_Check_QA}{\texttt{Code}}
  \faDatabase\ \href{https://hf.co/collections/sbhambr1/trace-check-qa-datasets}{\texttt{Datasets}}
}
\end{abstract}

\section{Introduction}
\label{sec:introduction}
Reasoning with intermediate Chain-of-Thought (CoT)-style traces (step-by-step outputs that models produce prior to an answer) has become one of the defining strategies for improving the performance of Large Language Models (LLMs) over a diverse range of problems, as exemplified by approaches like DeepSeek R1 \citep{guo2025deepseek}. While models such as DeepSeek R1 often produce extremely verbose unstructured responses even for simple problems \citep{kambhampati2025reasoning}, these reasoning traces are utilized both as inference aids and supervision signals in Knowledge Distillation (KD) when Supervised Fine-Tuning (SFT) smaller LLMs for enhanced task performance \citep{magister2022teaching,Shridhar2022DistillingRC,tian2025beyond}.

A common but often implicit assumption behind these CoT traces is that they are semantically correct and interpretable for end-users \citep{guo2025deepseek}. Training with these traces is done primarily to improve LLM performance on a given task, but fine-tuning objectives rarely require these traces to be semantically correct or interpretable. In this work, we challenge this assumption and ask: ``\textit{Must CoT reasoning traces be semantically correct and interpretable to end-user for enhancing LLM task performance?}"


\begin{figure}[ht]
    \centering
    \includegraphics[width=0.95\columnwidth]{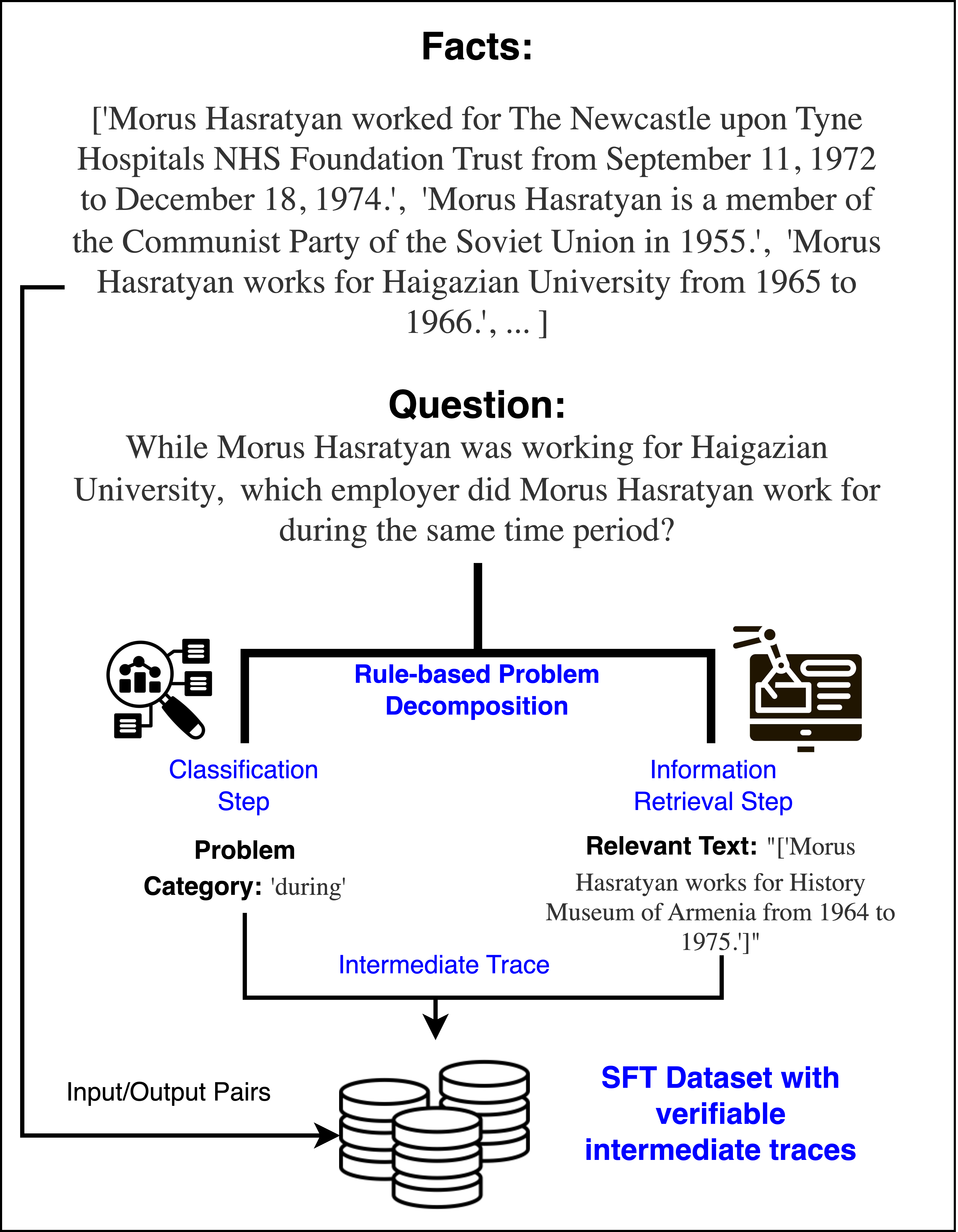}
    \caption{The construction of SFT dataset w/ verifiable intermediate traces using rule-based problem decomposition on an example from the CoTemp QA dataset.}
    \label{fig:example_fig}
\end{figure}

To address this, we focus our experiments on the Question Answering (QA) domain, where end-users regularly interact with both intermediate traces and final outputs \citep{polemi2024challenges} (e.g., ChatGPT \citep{chatgpt}, Perplexity \citep{perplexity}, Copilot \citep{microsoft_copilot}, Gemini \citep{google_gemini}). Faithfulness of reasoning traces is especially critical in these interactive settings where unverifiable traces can lead to loss of trust in users, misinformation and errors in model outputs, and perpetuation of biases among other negative consequences \citep{guidotti2018survey}. To assess the trade-offs between semantic correctness of the traces and LLM performance, we design an experimental setting where both final solutions and intermediate traces can be independently evaluated. Specifically, we employ a rule-based problem decomposition technique to break QA tasks into structured sub-problems \citep{mcdonald2024trace,xue2024decompose}. Next, we generate SFT datasets pairing questions with either verifiably correct or verifiably incorrect reasoning traces (while always including the correct answer). At inference time, this allows us to verify the correctness of both the final solution and the intermediate traces generated by the distilled model. 


To assess the trade-offs between \textcolor{black}{end-user interpretability} and LLM performance, we fine-tune models on different types of reasoning traces: DeepSeek R1 traces (verbose CoT outputs), LLM (GPT-4o-mini)-generated summaries of R1 traces (end-user facing summarizations), LLM (GPT-4o-mini)-generated post-hoc explanations (natural language explanations of R1 traces), and verifiably correct traces that we discussed above. In parallel, we conduct a human-subject study with 100 participants (hired on Prolific), split into four sets of 25. Each group was \textcolor{black}{asked to judge the interpretability} of the trace types using a Likert Scale measuring predictability, comprehensibility, and faithfulness attributes \citep{user-1, user-2}. 

Our experiments reveal two key findings: (1) \textbf{Correctness of CoT traces} is not reliably correlated with LLMs producing correct final answers: correct traces led to correct solutions only for 28\% test-set problems, while incorrect traces did not consistently degrade answer accuracy. (2) \textbf{End-user interpretability} of CoT traces is not reliably correlated with LLMs producing correct final answers:  fine-tuning on verbose DeepSeek R1 traces led to the strongest task performance, yet users rated these traces as least interpretable, scoring on average 3.39 for interpretability and 4.59 for cognitive load metrics on a 5-point Likert scale. These results highlight that \textbf{semantic correctness and human interpretability of reasoning traces can in fact be an albatross (around the neck of accuracy) from the perspective of LLM's task performance}, challenging assumptions in current LLM supervision practices.

The paper is organized as follows: \S \ref{sec:related_work} reviews prior work on Large Language \& Reasoning Models, Knowledge Distillation, and CoT Trace Interpretability. \S \ref{sec:method} presents our problem setup, rule-based decomposition for Open-Book QA, and dataset construction for distilling LLMs with correct and incorrect intermediate traces. \S \ref{sec:expts} describes the SFT experiments and human-subject studies, and \S \ref{sec:results} analyzes results and key insights. We conclude the work in \S \ref{sec:conclusion}. Appendix  includes details on datasets and prompts in App. \ref{sec:appendix_expts}, followed by user study and additional results in App. \ref{sec:appendix_user_study}.
\section{Related Work}
\label{sec:related_work}

\subsection{Large Reasoning Models \& CoT traces}
\label{subsec:related_work_llms}
Large Language Models (LLMs) have shown remarkable performance on a wide variety of natural language tasks in question answering, text generation, summarization, and translation, to name a few \citep{bubeck2023sparks}. Recent advances in post-training techniques have led to the rise of Large Reasoning Models (LRMs) such as DeepSeek R1 \citep{guo2025deepseek}, Google Gemini 2.5 \citep{google_gemini}, Microsoft Phi-4-reasoning \citep{abdin2025phi}, etc. These reasoning models produce a set of intermediate tokens, commonly referred to as `reasoning' traces, followed by the final solution. While LRMs have shown a significant improvement in final solution accuracy on reasoning tasks over standard LLMs \citep{guo2025deepseek, abdin2025phi}, their intermediate traces are meandering and verbose \textcolor{black}{\citep{reasoning3,reasoning2}}, making it hard to evaluate their trace validity and \textcolor{black}{end-user interpretability} \citep{kambhampati2025reasoning, kambhampati2025stop}.

\subsection{Knowledge Distillation}
\label{subsec:related_work_kd}
While Small Language Models (SLMs) offer a computationally efficient alternative to LLMs and LRMs, they are not robust to prompt augmentations (such as Chain-of-Thought) or steerable using in-context examples used in few-shot prompt settings \citep{Shridhar2022DistillingRC,Stolfo2022ACF}. Knowledge Distillation is a well-studied approach used for fine-tuning these SLMs (student) via the outputs of a larger model (teacher) \citep{magister2022teaching}. With LRMs generating both an intermediate trace and the final solution, SLMs are also distilled to replicate this output \citep{Shridhar2022DistillingRC,tian2025beyond}.  However, the lack of structured intermediate trace outputs makes the validity of the traces hard to evaluate \textcolor{black}{\citep{zhou2022least,chen2022program,guo2025deepseek}}. This problem is exacerbated for end-user settings such as in Question Answering (QA) domains, where user interactions involve exposure to both intermediate traces and final outputs.


\subsection{Interpretability of CoT Traces}
\label{subsec:related_works_interpretability}
Some recent works have argued for making these CoT traces more interpretable to the end-user, i.e., improve their faithfulness for the end user, as they are believed to serve as the LLM's explanation to generate the final solution \citep{arcuschin2025chain,tanneru2024hardness,li2024towards,tutek2025measuring,paul2024making,lyu2023faithful,lanham2023measuring,yeo2024interpretable}. On the other hand, there has also been work showcasing why these traces are not explainable to the end user \citep{barez2025chain}. Both sides of this argument stem from the assumption that these traces are indeed meant to be useful and interpretable for the end user and not just for the LLM to improve its final solution performance over a certain task. We specifically challenge this assumption and show the disconnect between the use of CoT traces for the LLM (as a training signal in SFT) and the use of CoT traces for the end user (as an interpretable reason behind the model's final solution).

\section{Knowledge Distillation using Problem Decomposition}
\label{sec:method}

This section describes our rule-based problem decomposition method for breaking complex Open Book QA tasks into verifiable sub-problems (\S \ref{subsec:method_formalization}) and explains how we use it to generate structured intermediate traces for SLM distillation (\S \ref{subsec:method_trace_gen}).

\subsection{Rule-based Problem Decomposition}
\label{subsec:method_formalization}

In the context of Open Book QA, consider the example shown in Figure \ref{fig:example_fig} which consists of a text passage (or a set of facts) and a question involving temporal reasoning between the queried problem and the facts present in the provided text. Answering this reasoning question involves identifying the relevant fact from the text which satisfies the temporal relation asked in the problem. In this case, the queried fact refers to \textit{``Morus Hasratyan works for Haigazian University from 1965 to 1966."} The temporal relation queried in the problem is `during' and thus, the relevant fact that answers the query is \textit{``Morus Hasratyan works for History Museum of Armenia from 1964 to 1975."} Hence, the final answer is \textit{`History Museum of Armenia'}. From this example, we see that the complex Open Book QA problem can be decomposed into a 1) \textbf{Classification step} determining the type of question asked (`during' temporal relation in this case), and an 2) \textbf{Information Retrieval (IR) step} to determine the relevant part of text that can answer the query. Therefore, we utilize these two steps to decompose the Open Book QA problems that allow us to construct structured intermediate traces for evaluation.

\subsection{Intermediate Trace Generation for SFT}
\label{subsec:method_trace_gen}

Given the outputs of the sub-problems obtained by decomposing the original query as shown in Figure \ref{fig:example_fig}, we generate the intermediate traces in an automated way which consists of the Classification step describing the type of the question posed in the query, and the IR step showing the relevant fact in the text that can help answer the query. We construct a dataset using these Input-Trace-Output tuples that can be utilized to SFT the Small Language Models. Note, that by constructing the intermediate trace using these two steps, we can then evaluate the accuracy of the intermediate traces generated by the distilled model at inference time. We will refer to this setting as \textbf{SFT w/ Correct Traces} for further discussion. 

To critically understand the correlation between intermediate trace correctness and final solution accuracy for Knowledge Distillation methods, we also consider an alternative SFT setting where for every input problem, we choose an incorrect problem category and incorrect fact/s for constructing the intermediate trace. This allows us to construct a SFT dataset which also consists of Input-Trace-Output tuples but with incorrect traces and correct final outputs. We will refer to this setting as \textbf{SFT w/ Incorrect Traces}.\footnote{An incorrect trace could also be generated by a noisy verifier such as an LLM-as-a-judge, but LLM-based verification is fundamentally noisy. Introducing this noise into the SFT dataset would make it impossible to determine if a student model’s performance was driven by the actual correctness of the trace or by the artifacts/biases of the noisy verifier. By using a rule-based problem decomposition method, we ensure that the intermediate traces are verifiably correct by construction. This allows for a binary, non-probabilistic evaluation of the reasoning steps (Classification and Information Retrieval).} 


\section{Experimental Setup}
\label{sec:expts}

This section presents the Open Book QA datasets used in our experiments (\S \ref{subsec:expts_datasets}), details our analysis of the link between trace semantic correctness and LLM performance (\S \ref{subsec:expts_semantics}), and describes the SFT experiments and human study evaluating interpretability–performance trade-offs (\S \ref{subsec:expts_interpretability}). Implementation details are described in \S \ref{subsec:expts_implementation}.


\subsection{Datasets}
\label{subsec:expts_datasets}
\paragraph{CoTemp QA:} CoTemp QA \citep{su2024living} is an English dataset of co-temporal questions requiring identification of a temporal relation type and inference of the corresponding fact in a passage. It includes four relation types—equal, overlap, during, and mix—and typically needs one or two supporting facts per question. We use 3,798 training and 950 test samples for our SFT experiments.


\paragraph{Microsoft MARCO QA:} The Microsoft MAchine Reading COmprehension (MARCO) dataset \citep{bajaj2016ms} is an English dataset of real user queries from Bing, each accompanied by long passages sourced from supporting URLs. It includes five query types—description, numeric, entity, location, and person—and typically requires one paragraph to answer. We use 5,000 training and 1,000 test samples in our experiments.


\paragraph{Facebook bAbI QA:} The Facebook bAbI QA dataset \citep{weston2015towards} is an English benchmark for evaluating reading comprehension through QA tasks that test reasoning skills such as fact chaining and deduction. While the full dataset contains 20 categories, we use 11 in our experiments—single-supporting-fact, two-supporting-facts, two-arg-relations, counting, lists-sets, conjunction, time-reasoning, basic-deduction, basic-induction, positional-reasoning, and size-reasoning. Each question typically requires about three supporting facts. Our SFT dataset includes 3,773 training and 376 test samples, with detailed splits provided in \S \ref{subsec:appendix_datasets}.


\begin{table*}[t]
\centering
\caption{An example from the CoTemp QA dataset showing the outputs of Qwen3-1.7B and Llama-3.2-1B-Instruct models under different query setting. Correct final solutions are shown in \textcolor{green}{green}, and incorrect final solutions are shown in \textcolor{red}{red}. Correct intermediate traces are shown in \textcolor{black}{blue}, and incorrect intermediate traces are shown in \textcolor{red}{red}.}
\label{tab:cotemp_example}
\resizebox{\textwidth}{!}{%
\begin{tabular}{@{}cc|c@{}}
\toprule
\textbf{Model} &
  \textbf{Query Setting} &
  \textbf{Example} \\ \midrule
 &
   &
  \cellcolor[HTML]{ECF4FF}\begin{tabular}[c]{@{}c@{}}\textbf{Input Prompt:} 'Answer the question based on the context: \\ {[}\textbackslash{}'Morus Hasratyan worked for The Newcastle upon Tyne Hospitals NHS Foundation Trust \\ from September 11, 1972 to December 18, 1974.\textbackslash{}', \\ \textbackslash{}'Morus Hasratyan is a member of the Communist Party of the Soviet Union in 1955.\textbackslash{}', \\ \textbackslash{}'Morus Hasratyan works for Haigazian University from 1965 to 1966.\textbackslash{}', "\\ "Morus Hasratyan worked for Bishop\textbackslash{}'s University from 1972 to 1975."", \\ \textbackslash{}'Morus Hasratyan worked for ISCTE – Lisbon University Institute from June, 1957 to December, 1960.\textbackslash{}', \\ \textbackslash{}'Morus Hasratyan works for History Museum of Armenia from 1964 to 1975.\textbackslash{}', \\ \textbackslash{}'Morus Hasratyan worked for Royal Air Force College Cranwell in February, 1959.\textbackslash{}', \\ \textbackslash{}'Morus Hasratyan worked for University of Detroit Mercy in September, 1963.\textbackslash{}', \\ \textbackslash{}'Morus Hasratyan worked for Tagesspiegel from May, 1957 to November, 1957.\textbackslash{}', \\ \textbackslash{}'Morus Hasratyan worked for North Carolina State University in May, 1962.\textbackslash{}'{]} \\ Question: While Morus Hasratyan was working for Haigazian University, \\ which employer did Morus Hasratyan work for during the same time period? Only return the answer.\end{tabular} \\ \cmidrule(l){3-3} 
 &
   &
  \cellcolor[HTML]{ECF4FF}\begin{tabular}[c]{@{}c@{}}\textbf{Gold Answer:} "{[}'History Museum of Armenia'{]}" ; \\ \textbf{Problem Category:} 'during' ; \\ \textbf{Relevant Text:} "{[}'Morus Hasratyan works for History Museum of Armenia from 1964 to 1975.'{]}"\end{tabular} \\ \midrule
\multirow{4}{*}{\textbf{Qwen3-1.7b}} &
  Prompt &
  "\textcolor{red}{Bishop's University}" \\ \cmidrule(l){3-3} 
 &
  SFT - Vanilla &
  \cellcolor[HTML]{EFEFEF}"\textcolor{green}{{[}'History Museum of Armenia'{]}}" \\ \cmidrule(l){3-3} 
 &
  SFT w/ Correct Traces &
  \begin{tabular}[c]{@{}c@{}}"\textless{}think\textgreater{}The temporal relation between the event in question and the event in context is: \textcolor{red}{equal}. \\ I need to use the following facts to answer the question: \\  \textcolor{black}{{[}'Morus Hasratyan works for History Museum of Armenia from 1964 to 1975.'{]}}\textless{}/think\textgreater \\ \textless{}answer\textgreater{}\textcolor{green}{{[}'History Museum of Armenia'{]}}\textless{}/answer\textgreater{}"\end{tabular} \\ \cmidrule(l){3-3} 
 &
  SFT w/ Incorrect Traces &
  \cellcolor[HTML]{EFEFEF}\begin{tabular}[c]{@{}c@{}}"\textless{}think\textgreater{}The temporal relation between the event in question and the event in context is:  \textcolor{red}{equal}. \\ I need to use the following facts to answer the question: \\  \textcolor{red}{Morus Hasratyan worked for ISCTE - Lisbon University Institute from June, 1957 to December, 1960.}\textless{}/think\textgreater \\ \textless{}answer\textgreater{}\textcolor{green}{{[}'History Museum of Armenia'{]}}\textless{}/answer\textgreater{}"\end{tabular} \\ \midrule
\multirow{4}{*}{\textbf{Llama-3.2-1B-It}} &
  Prompt &
  "\textcolor{red}{Bishop's University}" \\ \cmidrule(l){3-3} 
 &
  SFT - Vanilla &
  \cellcolor[HTML]{EFEFEF}"\textcolor{green}{{[}'History Museum of Armenia'{]}}" \\ \cmidrule(l){3-3} 
 &
  SFT w/ Correct Traces &
  \begin{tabular}[c]{@{}c@{}}"\textless{}think\textgreater{}The temporal relation between the event in question and the event in context is: \textcolor{black}{during}. \\ I need to use the following facts to answer the question: \\ \textcolor{black}{{[}'Morus Hasratyan works for History Museum of Armenia from 1964 to 1975.'{]}}\textless{}/think\textgreater \\ \textless{}answer\textgreater{}\textcolor{green}{{[}'History Museum of Armenia'{]}}\textless{}/answer\textgreater{}"\end{tabular} \\ \cmidrule(l){3-3} 
 &
  SFT w/ Incorrect Traces &
  \cellcolor[HTML]{EFEFEF}\begin{tabular}[c]{@{}c@{}}"\textless{}think\textgreater{}The temporal relation between the event in question and the event in context is: \textcolor{red}{overlap}. \\ I need to use the following facts to answer the  question: \\ \textcolor{red}{Morus Hasratyan worked for Royal Air Force College Cranwell in February, 1959.}\textless{}/think\textgreater \\ \textless{}answer\textgreater{}\textcolor{green}{{[}'History Museum of Armenia'{]}}\textless{}/answer\textgreater{}"\end{tabular} \\ \bottomrule
\end{tabular}%
}
\end{table*}
\subsection{Trace Accuracy vs Task Performance}
\label{subsec:expts_semantics}
For our experiments to evaluate the correlation between intermediate trace correctness and final solution accuracy, we utilize the Llama-3.2-1B-Instruct and the Qwen3-1.7B chat models. We adopt the following baselines for our evaluations:

\textbf{Direct Prompting SLMs:} We directly prompt the two SLMs to establish the baseline performance of these models across the three datasets without any additional fine-tuning.

\textbf{SFT - Vanilla:} Following the conventional fine-tuning technique, we also utilize the SFT baseline where we fine-tune the models using only Input-Output pairs and no intermediate traces. The input is same as the prompt used in the above setting. This allows us to evaluate the final solution performance for these models against the final solution performance obtained via directly prompting and via SFT with intermediate traces.

To examine intermediate trace correctness, we run the following experiments:

\textbf{SFT w/ Correct Traces:} Using the intermediate traces constructed via problem decomposition (Section \ref{subsec:method_trace_gen}), we fine-tune the models using the Input-Trace-Output tuples for each of the three datasets.

\textbf{SFT w/ Incorrect Traces:} In this case, we construct incorrect intermediate traces as discussed in Section \ref{subsec:method_trace_gen}, but use the correct final solutions in the Input-Trace-Output tuples.

For our experiments on SFT w/ Correct Traces and SFT w/ Incorrect Traces, we additionally report Category Accuracy (signifying the Classification step performance in the intermediate trace), the IR step Accuracy (signifying the IR step performance in the intermediate trace), and average trace length (\# of words in the intermediate trace) on the test datasets computed at inference time.

\subsection{Interpretability of Traces vs Final Solution Accuracy}
\label{subsec:expts_interpretability}
To assess the correlation between end-user interpretability of intermediate traces and final solution accuracy, we also evaluate larger models (Qwen3-8B and Llama-3.1-8B) to study the effects of SFT with more complex traces. Due to practical constraints of human-subject studies, these experiments are conducted only on the CoTemp QA domain (\S \ref{subsec:expts_datasets}).

\subsubsection{Reasoning Trace Generation} 
We consider (1) DeepSeek R1 traces where we prompt the R1 model on the CoTemp QA training dataset and collect the model responses for our SFT experiments where it got the correct final answer. Utilizing this filtered training dataset, we prompt GPT-4o-mini to generate both (2) summaries and (3) post-hoc explanations of these R1 traces. Since R1 traces can often be verbose, we posit that their summary as well as a post-hoc explanation can likely be more interpretable to the end user (prompts shown in \S \ref{sec:appendix_prompts}).

\begin{table*}[t]
\centering
\small
\caption{CoTemp QA Results}
\label{tab:cotempqa_results}
\resizebox{\textwidth}{!}{%
\begin{tabular}{@{}cc|cccc|ccc@{}}
\toprule
 &
   &
  \multicolumn{4}{c|}{\textbf{Final Solution Evaluations}} &
  \multicolumn{3}{c}{\textbf{Intermediate Trace Evaluations}} \\ \midrule
\textbf{Model} &
  \textbf{Query Setting} &
  \textbf{Accuracy} &
  \textbf{F1} &
  \textbf{Precision} &
  \textbf{Recall} &
  \begin{tabular}[c]{@{}c@{}}\textbf{Classification} \\ \textbf{Step Accuracy}\end{tabular} &
  \begin{tabular}[c]{@{}c@{}}\textbf{IR Step} \\ \textbf{Accuracy}\end{tabular} &
  \begin{tabular}[c]{@{}c@{}}\textbf{Avg Trace} \\ \textbf{Length} (\# tokens)\end{tabular} \\ \midrule
\multirow{4}{*}{\textbf{Qwen3-1.7b}}   & Prompt                & 6.35   & 11.35  & 14.33 & 10.1   & -     & -     & -     \\
                              & SFT - Vanilla         & 60.33 & 74.88 & 82.15  & 71.3   & -     & -     & -     \\
                              & SFT - Correct Trace   & 52.88 & 70.63 & 79.45  & 66.33 & \textbf{47.06} & \textbf{78.99} & 45.8  \\
                              & SFT - Incorrect Trace & \textbf{63.88} & \textbf{76.5}   & \textbf{82.58} & \textbf{73.5 }  & 20.36 & 56.92 & 34.15 \\ \midrule
\multirow{4}{*}{\textbf{Llama-3.2-1B-It}} & Prompt                & 7.48  & 13.78 & 17.58 & 12.15  & -     & -     & -     \\
                              & SFT - Vanilla         & 44.65  & 61.08 & 69.53 & 56.58 & -     & -     & -     \\
                              & SFT - Correct Trace   & 39.55  & 56.83 & 65.83 & 52.5   & \textbf{39.09} & \textbf{79.4}  & 43.51 \\
                              & SFT - Incorrect Trace & \textbf{45.58} & \textbf{61.15}  & \textbf{69.65}  & \textbf{57.23} & 18.8  & 73.62 & 40.28 \\ \bottomrule
\end{tabular}%
}
\end{table*}
\begin{table*}[t]
\centering
\small
\caption{Microsoft MARCO QA and Facebook bAbI QA Results}
\label{tab:marco_babi_results}
\resizebox{\textwidth}{!}{%
\begin{tabular}{cc|cccc|cccc}
\toprule
\textbf{} &
  \textbf{} &
  \multicolumn{4}{c|}{\textbf{Microsoft MARCO QA}} &
  \multicolumn{4}{c}{\textbf{Facebook bAbI QA}} \\ \midrule
\textbf{Model} &
  \textbf{Query Setting} &
  \multicolumn{1}{c|}{\textbf{\begin{tabular}[c]{@{}c@{}}Avg Final \\ Solution \\ Accuracy (\%)\end{tabular}}} &
  \textbf{\begin{tabular}[c]{@{}c@{}}Avg Trace Acc \\ (Classification \\ Step) (\%)\end{tabular}} &
  \textbf{\begin{tabular}[c]{@{}c@{}}Avg Trace \\ Acc (IR \\ Step) (\%)\end{tabular}} &
  \textbf{\begin{tabular}[c]{@{}c@{}}Avg Trace \\ Length\\ (\# tokens)\end{tabular}} &
  \multicolumn{1}{c|}{\textbf{\begin{tabular}[c]{@{}c@{}}Avg Final \\ Solution \\ Accuracy (\%)\end{tabular}}} &
  \textbf{\begin{tabular}[c]{@{}c@{}}Avg Trace Acc \\ (Classification \\ Step) (\%)\end{tabular}} &
  \textbf{\begin{tabular}[c]{@{}c@{}}Avg Trace \\ Acc (IR \\ Step) (\%)\end{tabular}} &
  \textbf{\begin{tabular}[c]{@{}c@{}}Avg Trace \\ Length \\ (\# tokens)\end{tabular}} \\ \midrule
\multirow{4}{*}{\textbf{Qwen3-1.7B}} &
  Prompt &
  \multicolumn{1}{c|}{0} &
  - &
  - &
  - &
  \multicolumn{1}{c|}{0} &
  - &
  - &
  - \\
 &
  SFT - Vanilla &
  \multicolumn{1}{c|}{3.4} &
  - &
  - &
  - &
  \multicolumn{1}{c|}{\textbf{97.9}} &
  - &
  - &
  - \\
 &
  SFT - Correct Trace &
  \multicolumn{1}{c|}{\textbf{26.3}} &
  \textbf{60.4} &
  40.6 &
  68.14 &
  \multicolumn{1}{c|}{94.41} &
  \textbf{60.64} &
  \textbf{24.73} &
  43.25 \\
 &
  SFT - Incorrect Trace &
  \multicolumn{1}{c|}{20.3} &
  6.9 &
  \textbf{52.5} &
  85.07 &
  \multicolumn{1}{c|}{95.21} &
  17.82 &
  0 &
  42.45 \\ \midrule
\multirow{4}{*}{\textbf{Llama-3.2-1B-It}} &
  Prompt &
  \multicolumn{1}{c|}{1.7} &
  - &
  - &
  - &
  \multicolumn{1}{c|}{12.8} &
  - &
  - &
  - \\
 &
  SFT - Vanilla &
  \multicolumn{1}{c|}{33.4} &
  - &
  - &
  - &
  \multicolumn{1}{c|}{\textbf{96.5}} &
  - &
  - &
  - \\
 &
  SFT - Correct Trace &
  \multicolumn{1}{c|}{\textbf{33.7}} &
  \textbf{59.9} &
  21.4 &
  55.82 &
  \multicolumn{1}{c|}{94.41} &
  \textbf{61.7} &
  \textbf{24.73} &
  42.17 \\
 &
  SFT - Incorrect Trace &
  \multicolumn{1}{c|}{28.9} &
  20 &
  \textbf{43.9} &
  80.48 &
  \multicolumn{1}{c|}{86.17} &
  3.46 &
  0 &
  38.5 \\ \bottomrule
\end{tabular}%
}
\end{table*}

\subsubsection{Human-Subject Study}
We conducted four separate user studies to evaluate the \textcolor{black}{end-user interpretability} of the four types of reasoning traces. In each study, a set of 25 participants were hired on Prolific and shown only one type of trace: (1) DeepSeek R1 traces, (2) summarized R1 traces, (3) post-hoc explanations of R1 traces, or (4) verifiably correct reasoning traces. An example of the four types of traces can be found in \S\ref{sec:app_user_traces}. We use a between-subjects design to avoid bias from having participants compare multiple trace types themselves. We specifically test the following hypotheses:

\textbf{H1:} Reasoning traces that improve task accuracy will not lead to higher interpretability for the user.

\textbf{H2:} Reasoning traces that improve task accuracy will be associated with higher cognitive workload for the user, as measured by increased mental demand, effort, and frustration.

Each participant evaluated five fixed Q/A examples containing the question, predicted answer, and reasoning trace. After each example, participants rated the trace on a 5-point Likert scale for predictability, comprehensibility, interpretability, and faithfulness to context \citep{user-2, user-1}. To assess cognitive workload, we employed the NASA–TLX \citep{hart2006nasa}, focusing on mental demand, effort, and frustration. Further details on the user study, participant demographics, and procedures are provided in \S \ref{sec:appendix_user_study}.


\subsection{Implementation Details}
\label{subsec:expts_implementation}
Models were fine-tuned using the Hugging Face library \citep{wolf-etal-2020-transformers} on a single 80GB NVIDIA Tesla A100 GPU for 3 epochs (effective batch size 16, max sequence length 1024). We employed PEFT QLoRA \citep{dettmers2023qlora} (rank 16, alpha 32) with a learning rate of 2e-4 (8-bit AdamW, cosine scheduler, 0.1 warm-up). Prompt experiments utilized vLLM \citep{kwon2023efficient}.

\section{Results}
\label{sec:results}

\subsection{Final Solution \& CoT Performance}
\label{subsec:results_intermediate_accuracy}
As shown in Table \ref{tab:cotempqa_results}, SFT with incorrect traces yields the highest final solution scores across Accuracy, Precision, F1, and Recall for both models. However, models SFT-ed with correct traces achieve higher Classification and IR Step accuracies due to the verifiability of their intermediate traces. Consistent patterns appear in the MARCO QA and bAbI QA datasets (Table \ref{tab:marco_babi_results}): while final solution accuracy remains comparable, models trained with correct traces show stronger intermediate trace performance—particularly in bAbI QA, where those trained with incorrect traces perform poorly on the trace accuracy.

\begin{figure}[tp]
    \centering

    \begin{subfigure}[b]{0.48\columnwidth}
        \includegraphics[width=\textwidth]{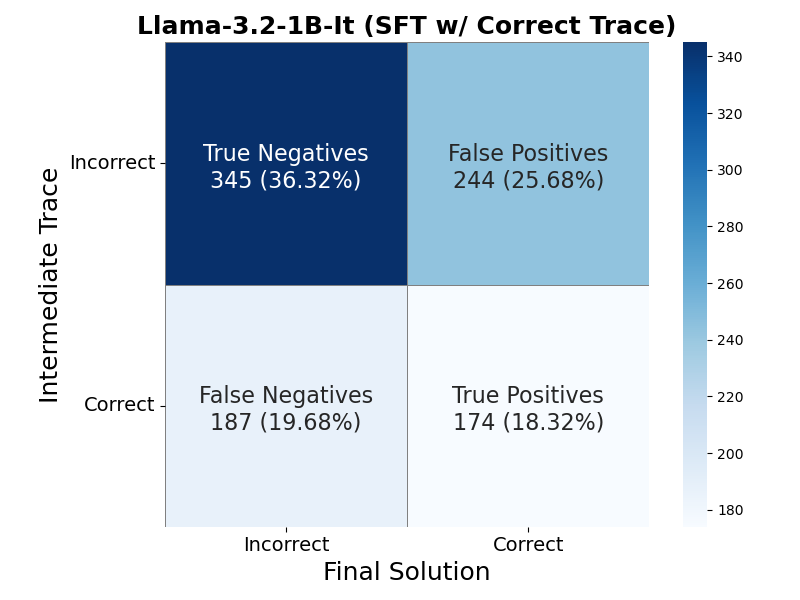}
        \caption*{\small CoTemp (Llama, Correct)}
    \end{subfigure}\hfill%
    \begin{subfigure}[b]{0.48\columnwidth}
        \includegraphics[width=\textwidth]{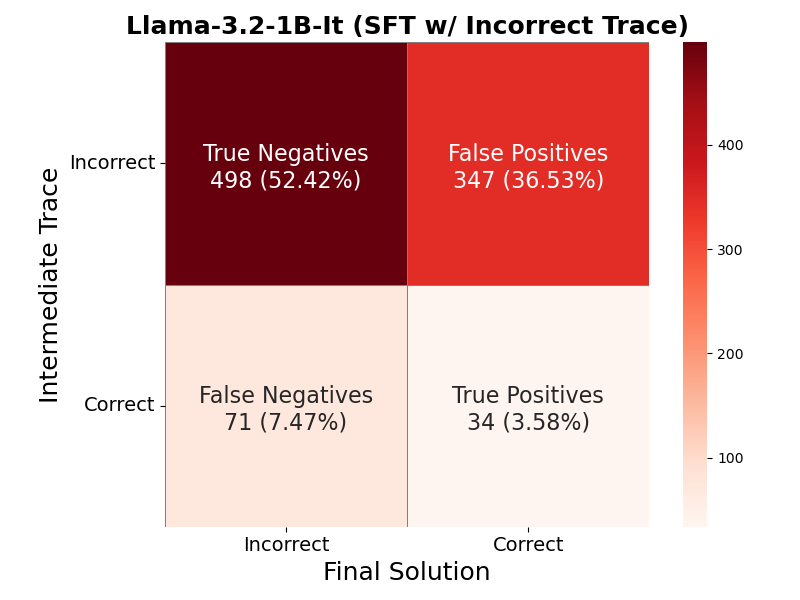}
        \caption*{\small CoTemp (Llama, Incorrect)}
    \end{subfigure}

    \vspace{0.3em}

    \begin{subfigure}[b]{0.48\columnwidth}
        \includegraphics[width=\textwidth]{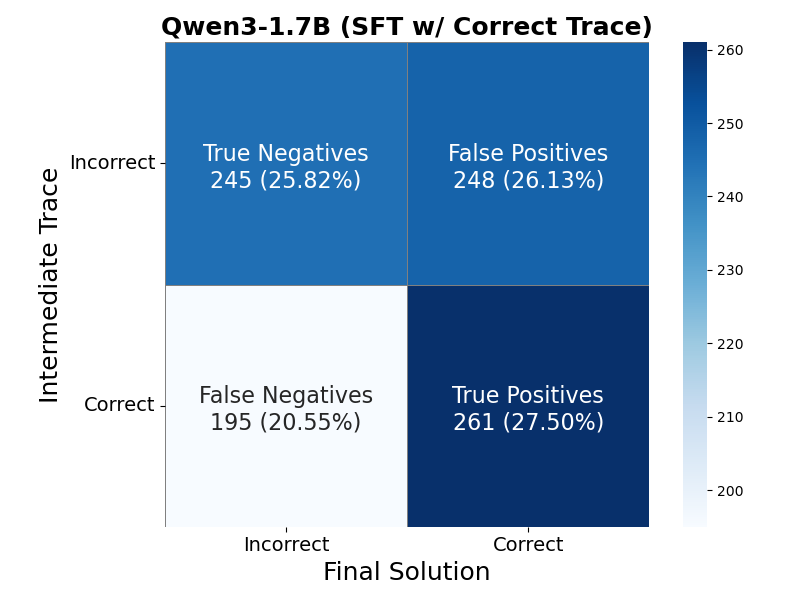}
        \caption*{\small CoTemp (Qwen, Correct)}
    \end{subfigure}\hfill%
    \begin{subfigure}[b]{0.48\columnwidth}
        \includegraphics[width=\textwidth]{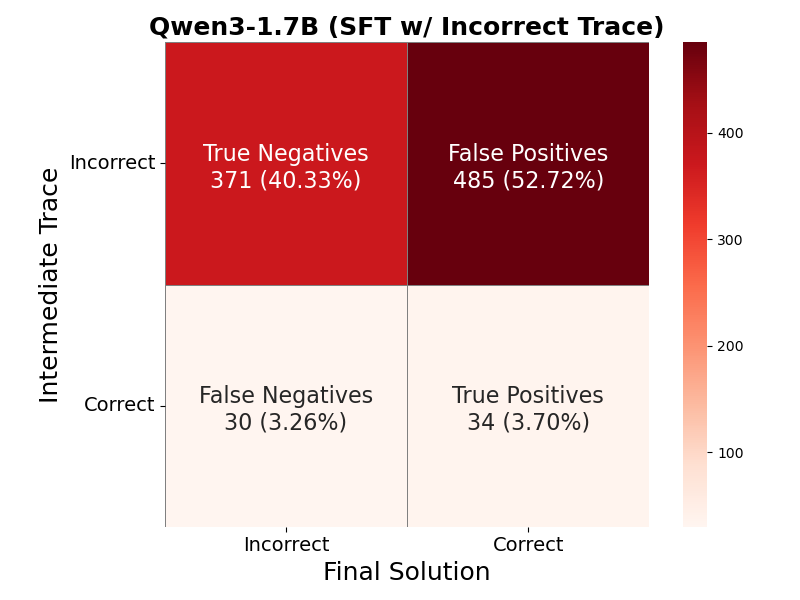}
        \caption*{\small CoTemp (Qwen, Incorrect)}
    \end{subfigure}

    \vspace{1.5em}

    \begin{subfigure}[b]{0.48\columnwidth}
        \includegraphics[width=\textwidth]{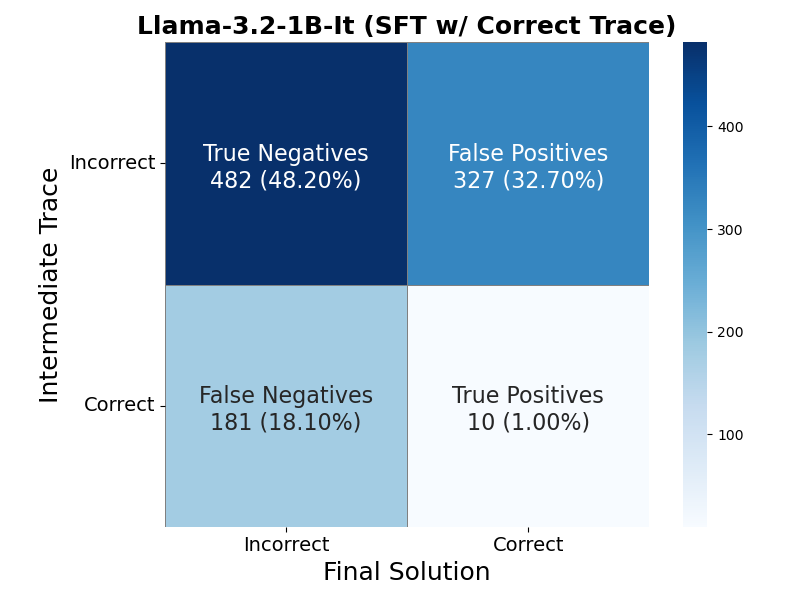}
        \caption*{\small MARCO (Llama, Correct)}
    \end{subfigure}\hfill%
    \begin{subfigure}[b]{0.48\columnwidth}
        \includegraphics[width=\textwidth]{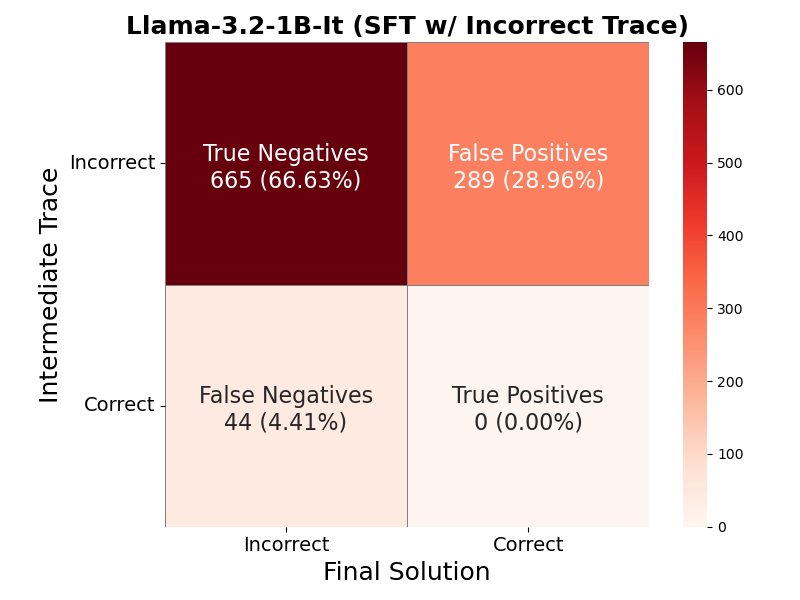}
        \caption*{\small MARCO (Llama, Incorrect)}
    \end{subfigure}

    \vspace{0.3em}

    \begin{subfigure}[b]{0.48\columnwidth}
        \includegraphics[width=\textwidth]{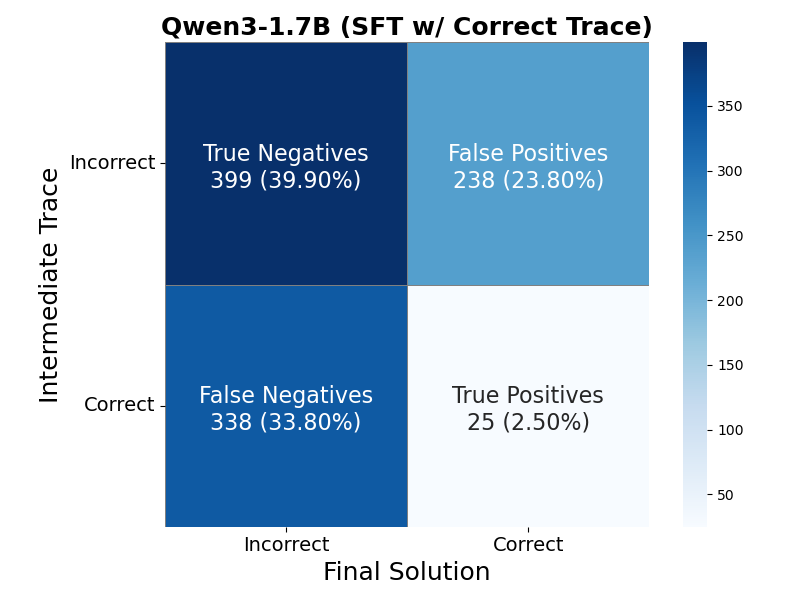}
        \caption*{\small MARCO (Qwen, Correct)}
    \end{subfigure}\hfill%
    \begin{subfigure}[b]{0.48\columnwidth}
        \includegraphics[width=\textwidth]{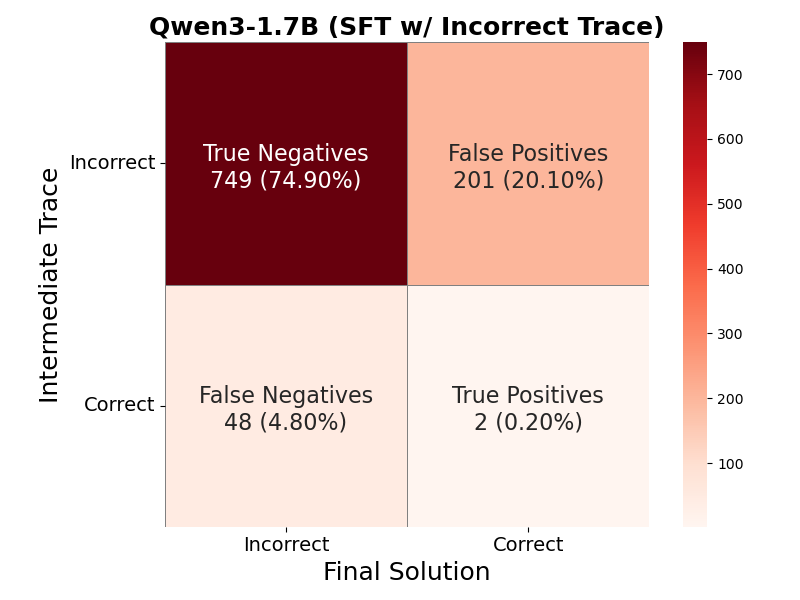}
        \caption*{\small MARCO (Qwen, Incorrect)}
    \end{subfigure}

    \vspace{1.5em}

    \begin{subfigure}[b]{0.48\columnwidth}
        \includegraphics[width=\textwidth]{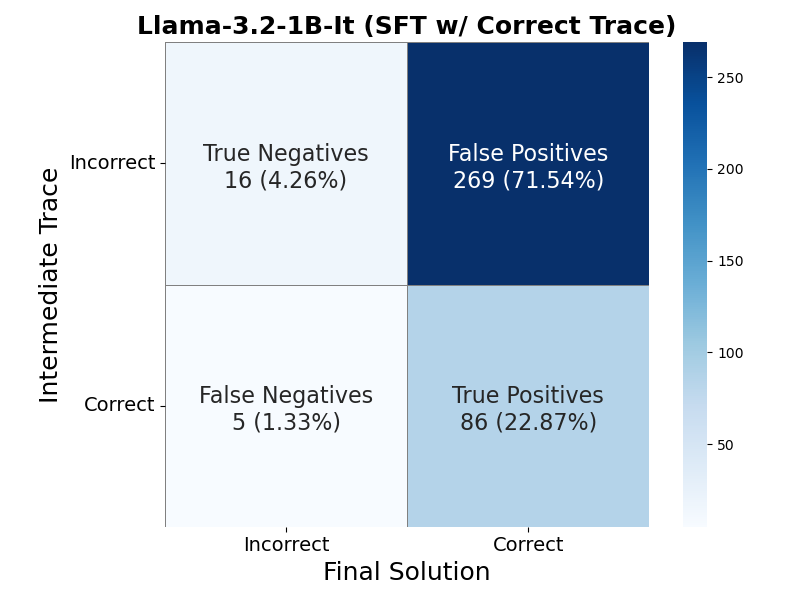}
        \caption*{\small bAbI (Llama, Correct)}
    \end{subfigure}\hfill%
    \begin{subfigure}[b]{0.48\columnwidth}
        \includegraphics[width=\textwidth]{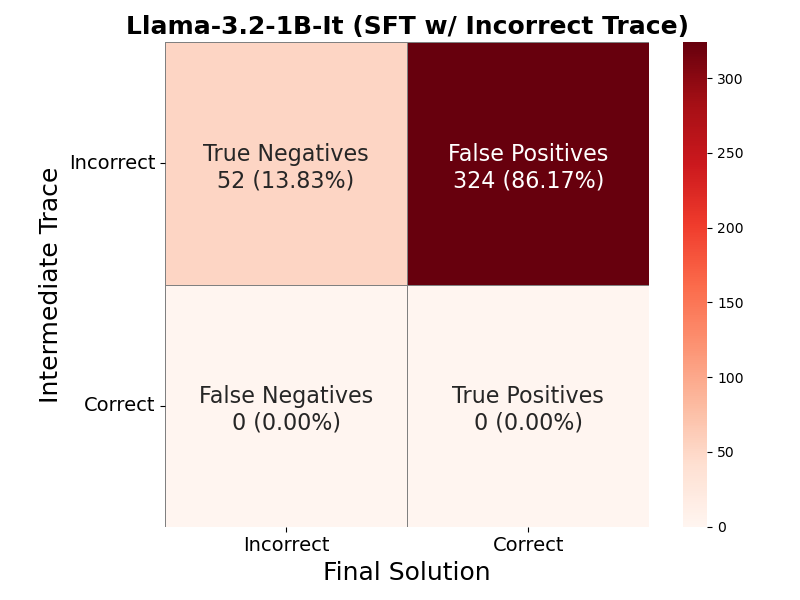}
        \caption*{\small bAbI (Llama, Incorrect)}
    \end{subfigure}

    \vspace{0.3em}

    \begin{subfigure}[b]{0.48\columnwidth}
        \includegraphics[width=\textwidth]{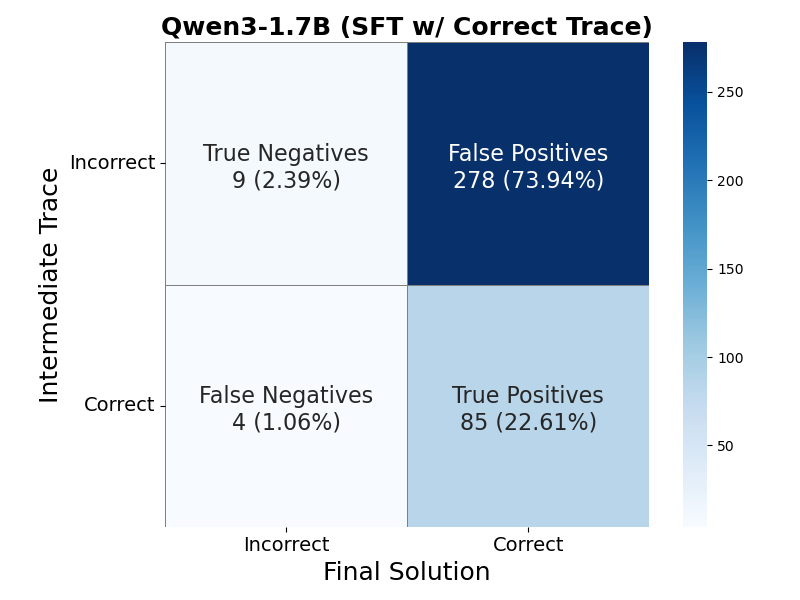}
        \caption*{\small bAbI (Qwen, Correct)}
    \end{subfigure}\hfill%
    \begin{subfigure}[b]{0.48\columnwidth}
        \includegraphics[width=\textwidth]{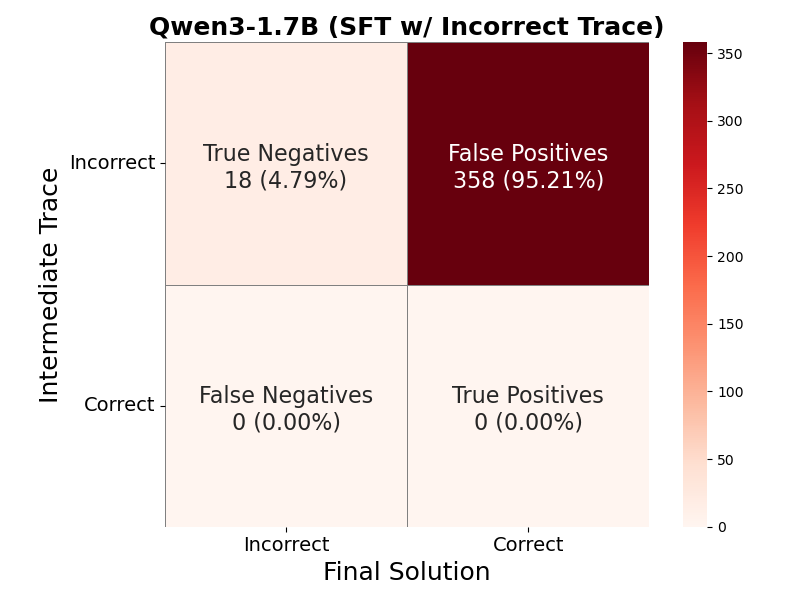}
        \caption*{\small bAbI (Qwen, Incorrect)}
    \end{subfigure}

    \caption{Confusion matrices showing Final Solution Accuracy (X-axis) vs Trace Accuracy (Y-axis) for the CoTemp QA, Microsoft MARCO QA, and Facebook bAbI QA datasets. Each row pairs Correct (left) vs Incorrect (right) traces for a given dataset--model combination.}
    \label{fig:sft_correct_incorrect_combined}
\end{figure}

\textcolor{black}{We hypothesize that the performance gain from incorrect traces arises from the model learning structural trace patterns while ignoring semantics. In our controlled setup, correct and incorrect traces share identical structures, and the SFT training with cross-entropy loss likely encourages the model to reproduce the paired <incorrect trace, correct final answer> during inference. We expect this finding to hold even when wrong traces contain only a flipped label or a single incorrect fact.}

\begin{table}[b]
\centering
\caption{Results on CoTemp QA with Llama-3.1-8B.}
\label{tab:llama8b}
\small
\begin{tabularx}{\columnwidth}{@{}X|c|cc@{}}\toprule
\multicolumn{1}{c|}{\textbf{Query Setting}} & \multicolumn{1}{c|}{\textbf{Final}} & \multicolumn{2}{c}{\textbf{Trace Accuracy}} \\
\cmidrule(l){3-4}
\multicolumn{1}{c|}{} & \multicolumn{1}{c|}{\textbf{Accuracy}} & \textbf{Cls. Step} & \textbf{IR Step} \\
\midrule
SFT - Vanilla            & \textbf{82.88} & ---   & ---   \\
SFT - Correct Trace      & 82.02 & \textbf{60.03} & \textbf{81.15} \\
SFT - Incorrect Trace    & 79.30 & 21.57 & 75.46 \\
\bottomrule
\end{tabularx}
\end{table}

\subsection{(Lack of) Correlation b/w Final Solution Accuracy \& Trace Correctness}
\label{subsec:results_semantic_correlation}

From the confusion matrices in Figure \ref{fig:sft_correct_incorrect_combined}, the top row (Llama-3.2-1B-It) and bottom row (Qwen3-1.7B) both show a high rate of False Positives—cases where models produce correct answers but incorrect traces (25.7\% in CoTemp, 32.7\% in MARCO, and 71.54\% in bAbI). In contrast, True Positives—both correct answers and traces—are few (e.g., 2.5\% in MARCO and 22.61\% in bAbI), indicating that many correct outputs rely on flawed intermediate traces. Figure \ref{fig:sft_correct_incorrect_combined} bottom row shows consistently high False Positive rates for both Llama-3.2-1B-It (top row) and Qwen3-1.7B (bottom row) across all datasets, indicating that finetuning on correct answers but incorrect traces enables strong final answer performance despite poor intermediate trace accuracy.

\begin{table*}[t]
\centering
\small
\renewcommand{\arraystretch}{1.05}
\setlength{\tabcolsep}{3pt}
\caption{Median participant ratings of reasoning traces across dimensions of \textcolor{black}{end-user interpretability} and cognitive workload. Arrows indicate the desired direction of scores: ↑ higher ratings are better for interpretability measures, ↓ lower ratings are better for cognitive workload measures.}
\label{tab:user-study}
\resizebox{\textwidth}{!}{%
\begin{tabular}{p{0.15\textwidth} | p{0.45\textwidth} | 
    >{\centering\arraybackslash}p{0.08\textwidth} 
    >{\centering\arraybackslash}p{0.10\textwidth} 
    >{\centering\arraybackslash}p{0.10\textwidth} 
    >{\centering\arraybackslash}p{0.08\textwidth}}
\toprule
\textbf{Dimension} & \textbf{Question} & 
\textbf{R1 Traces} & 
\textbf{Summarized R1} & 
\textbf{R1 Explanations} & 
\textbf{Correct Traces} \\
\midrule
Predictability & I could anticipate the next steps or conclusions based on earlier parts of the reasoning. (↑) & \textbf{3.48} & 4.45 & 4.29 & 4.82 \\
Comprehensibility & I understood the reasoning followed by the model. (↑) & \textbf{3.46} & 4.55 & 4.27 & 4.56 \\
 & I could follow each step in the reasoning without confusion. (↑) & \textbf{3.46} & 4.54 & 4.28 & 4.84 \\
Interpretability & The reasoning helped me understand why the model acted or concluded the way it did. (↑) & \textbf{3.31} & 4.53 & 4.29 & 4.86 \\
Faithfulness & There were no major gaps or missing reasoning steps in the reasoning. (↑) & \textbf{3.33} & 4.54 & 4.26 & 4.72 \\
 & The reasoning is consistent with the facts or evidence provided in the context. (↑) & \textbf{3.34} & 4.24 & 4.29 & 4.84 \\
\midrule
Mental Demand & How mentally demanding was the task? (↓) & \textbf{4.65} & 2.87 & 2.92 & 2.31 \\
Effort & How hard did you have to work to accomplish your level of performance? (↓) & \textbf{4.54} & 2.39 & 2.17 & 2.86 \\
Frustration & How frustrated, stressed, and annoyed were you? (↓) & \textbf{4.58} & 2.04 & 2.42 & 2.42 \\
\bottomrule
\end{tabular}
}
\end{table*}

Tables 2 and 3 show varied outcomes across datasets: SFT with incorrect traces performs best on CoTemp QA, SFT with correct traces leads on MARCO QA, and SFT-Vanilla outperforms both on bAbI QA. These results collectively refute the assumption that semantically correct traces necessarily improve final solution performance.

To further strengthen our argument, we have also conducted a {$\chi^2$} statistical test with the null hypothesis: Trace accuracy and answer accuracy are independent. The test was done for both Qwen3-1.7B and Llama-3.2-1B-Instruct for the SFT w/ Incorrect trace setting. With a degree of freedom of 1, alpha=0.05, the critical value is 3.841 and we obtained {$\chi^2$} to be 0.34 and 2.93 (both < 3.841) for the two models respectively. Hence, the null hypothesis is correct.

We further validate our findings, by extending our evaluation to a larger student model, Llama-3.1-8B, on the CoTemp QA dataset. As shown in Table~\ref{tab:llama8b}, results are consistent with our main findings: SFT with correct traces yields competitive final solution accuracy while maintaining higher trace accuracy, whereas SFT with incorrect traces degrades both trace classification and final solution accuracy. This confirms that our conclusions generalize beyond the smaller models evaluated in the main experiments.

\subsection{Error Analysis between Final Solution \& Intermediate Traces}
\label{subsec:results_error_analysis}

Figure \ref{fig:sft_correct_incorrect_combined} shows that even when models were SFT-ed with correct traces and solutions, a substantial portion of cases featured correct traces preceding incorrect final answers. For Llama-3.2-1B-It, this occurred in up to 51.8\% of CoTemp, 94.76\% of MARCO, and 5.5\% of bAbI samples; for Qwen3-1.7B, similar trends appeared (42.76\%, 93.11\%, and 4.49\%, respectively). These findings indicate that training with correct intermediate traces does not reliably yield correct final predictions across datasets.



\subsection{(Lack of) Correlation between Solution Accuracy \& Trace Interpretability}
\label{subsec:results_interpretability_correlation}

\subsubsection{SFT Evaluations} 
\begin{figure}[b]
    \centering
    \includegraphics[width=0.49\textwidth]{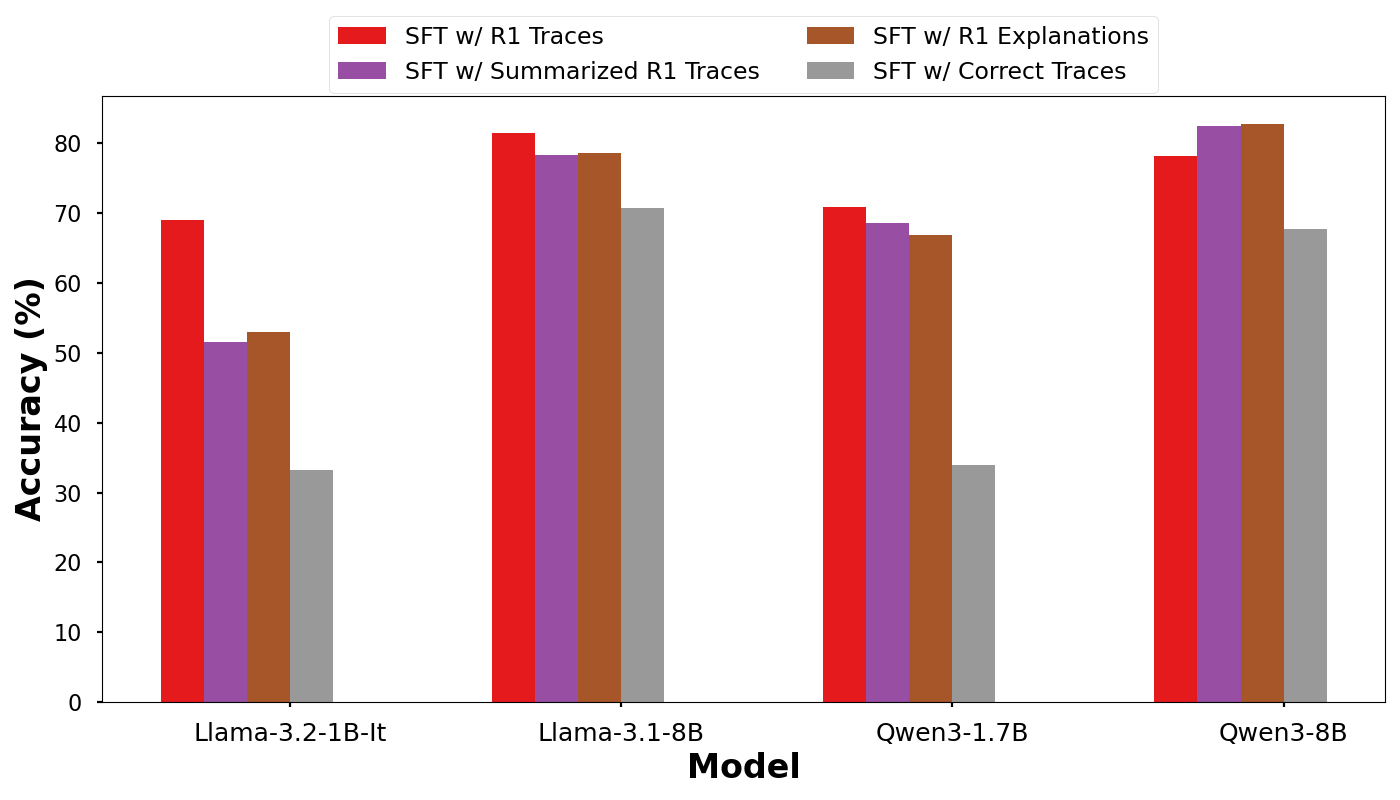}
    \caption{Final solution performance on CoTemp QA test dataset after SFT with different trace types on Llama and Qwen models.}
    \label{fig:trace_sft_comparison}
\end{figure}

Figure \ref{fig:trace_sft_comparison} shows that, except for Qwen3-8B, SFT with R1 traces achieves the highest final accuracy, with the largest gain in Llama-3.2-1B-Instruct. In contrast, models trained with algorithmically generated semantically verifiable (correct) traces perform worst, even compared to those using R1 summaries or explanations. These results motivate a user study to assess the human interpretability of R1 traces.

\subsubsection{Human-Subject Study Results} 
Table~\ref{tab:user-study} shows that participants found algorithmically generated correct traces most interpretable across all dimensions, while R1 traces scored lowest. Summarized and explained R1 traces received moderate ratings, suggesting improved comprehension by the subjects. R1 traces also imposed higher cognitive load, whereas correct traces caused less mental demand, effort, and frustration, making them easier to follow.

\subsubsection{Statistical Analysis}
We ran pairwise Mann–Whitney U tests with Bonferroni correction focusing on R1 vs. algorithmically generated correct traces for hypothesis testing with the following null hypotheses: \textbf{NH-1 (Interpretability):} There is no difference in interpretability ratings between R1 traces and algorithmically-generated correct reasoning traces. \textbf{NH-2 (Cognitive Workload):} There is no difference in cognitive workload ratings between R1 traces and algorithmically-generated correct reasoning traces.

Results show significant differences in both interpretability (predictability, comprehensibility, interpretability, faithfulness; all ) and cognitive workload (mental demand, effort, frustration; all ), leading us to reject both null hypotheses. Across all comparisons, R1 traces were consistently less interpretable and imposed higher cognitive load than other trace types.

\subsection{Discussion}
\label{subsec:discussion}

The key takeaways from our results can be summarized as follows:

1) \textit{SFT w/ incorrect traces at times outperformed SFT w/ correct traces in final solution accuracy (\S \ref{subsec:results_intermediate_accuracy}).}

2) \textit{Trace correctness did not guarantee final solution correctness. Solution correctness also did not imply a correct intermediate trace (\S \ref{subsec:results_semantic_correlation} \& \S \ref{subsec:results_error_analysis}).}

3) \textit{Fine-tuning LLMs with the traces found to be the least interpretable by end-users led to the highest final solution accuracy, and vice-versa (\S \ref{subsec:results_interpretability_correlation}).}

\section{Conclusion}
\label{sec:conclusion}

This work examines how an LLM’s final solution accuracy relates to the semantic correctness and end-user interpretability of its intermediate traces after SFT. Using a knowledge distillation approach with verifiable problem decomposition on Open Book QA tasks, we find little correlation between trace and solution accuracy across Llama and Qwen models. Although SFT with R1 traces yields the best performance, user studies show these traces are the least interpretable, imposing higher cognitive load. This decoupling suggests that (1) verbose traces aid model reasoning more than human understanding, and (2) generating user-friendly explanations requires separate objectives or modules. Overall, we call for trace-based training methods that better balance LLM performance with human interpretability.

While we focused on pretrained models fine-tuned on  Q\&A scenarios in this work, in companion work from our group, we have also investigated the end-user semantics of intermediate tokens starting from empty transformer models. Specifically, in \cite{stechly2025beyond}, we show that empty transformers trained on different distributions of maze problems, with the A* search traces as the verifiable intermediate tokens, exhibit  behavior very similar to what we reported here, including (i) loose correlation between solution accuracy and trace validity and (ii) robustness of accuracy with respect to training with swapped traces. Those studies also further show  (iii) lack of correlation between the length of intermediate tokens produced at inference time and the computational complexity of the problem instance and (iv) robustness of all these findings even in the face of RL-based post-training methods. In light of these findings, we have also written a position paper \cite{kambhampati2025stop} arguing against the anthropomorphization of intermediate tokens as ``reasoning traces."

\section*{Limitations}

OpenBook QA domain is not only representative of a critical application of today's interactive dialogue systems, but more importantly offers a controlled experiment environment for demonstrating the disconnect between trace fidelity and final accuracy. In other reasoning areas, as shown by commonly used test datasets like AIME/MATH500 for math and LiveCodeBench for code, there are many problems where creating a step-by-step reasoning process that works for every math or code problem is not possible. This creates a problem for evaluating the intermediate steps in these situations. The main point of this work is to demonstrate a gap between the accuracy of the steps and the final result for OpenBook QA problems, and the findings of this work are currently restricted to the aforementioned domains. These findings are also currently restricted by the choice and size of LLMs used in this work, and further efforts need to be carried to evaluate the extensibility of these results.

\section*{Acknowledgments}
This research is supported in part by grants from ONR (N00014-25-1-2301 and N00014-23-1-2409), DARPA (HR00112520016), DoD RAI (via CMU subcontract 25-00306-SUB-000), and a generous gift from Qualcomm. We also thank the ACL/ARR Area Chair for their thoughtful championing of our work. 

\bibliography{custom}
\appendix

\section{Additional Experiment Details}
\label{sec:appendix_expts}

\subsection{Dataset Distributions}
\label{subsec:appendix_datasets}

\begin{table}[ht]
\centering
\caption{Train and Test data distribution for CoTemp QA dataset used in our SFT experiments.}
\label{tab:cotemp_data}
\begin{tabular}{@{}c|c@{}}
\textbf{Category} & \textbf{Train/Test Samples} \\ \midrule
equal             & 349 / 87                    \\
overlap           & 522 / 131                   \\
during            & 2477 / 619                  \\
mix               & 450 / 113                   \\ \bottomrule
\end{tabular}%
\end{table}

\begin{table}[ht]
\centering
\caption{Train and Test data distribution for Microsoft MARCO QA dataset used in our SFT experiments.}
\label{tab:marco_data}
\begin{tabular}{@{}c|c@{}}
\textbf{Category} & \textbf{Train/Test Samples} \\ \midrule
description         & 1000 / 200                  \\
entity              & 1000 / 200                  \\
numeric             & 1000 / 200                  \\
location            & 1000 / 200                  \\ 
person              & 1000 / 200                  \\
\bottomrule
\end{tabular}%
\end{table}

\begin{table}[ht]
\centering
\caption{Train and Test data distribution for Facebook bAbI QA dataset used in our SFT experiments.}
\label{tab:babi_data}
\begin{tabular}{@{}c|c@{}}
\textbf{Category} & \textbf{Train/Test Samples} \\ \midrule
single-supporting-fact             & 200 / 20                    \\
two-supporting-facts               & 200 / 20                    \\
two-arg-relations                  & 1000 / 100                  \\
counting                           & 200 / 20                    \\ 
list-sets                          & 200 / 20                    \\ 
conjunction                        & 200 / 20                    \\ 
time-reasoning                     & 200 / 20                    \\ 
basic-deduction                    & 250 / 25                    \\ 
basic-induction                    & 1000 / 100                  \\ 
positional-reasoning               & 125 / 12                    \\ 
size-reasoning                     & 198 / 19                    \\ \bottomrule
\end{tabular}%
\end{table}

\subsection{Prompts}
\label{sec:appendix_prompts}
\begin{tcolorbox}[
    width=\columnwidth,
    colback=gray!5!white, 
    colframe=black, 
    title=R1 Trace Summarization Prompt, 
    fonttitle=\bfseries\large, 
    sharp corners, 
    boxrule=0.8pt,
    left=8pt, right=8pt, top=8pt, bottom=8pt
]
\large
Summarize the following trace in a very concise and clear manner, highlighting key events and outcomes in less than 100 words:

\vspace{4pt}
\textbf{\{R1 trace\}}

\vspace{4pt}
\textbf{Summary:}
\end{tcolorbox}

\vspace{10pt}

\begin{tcolorbox}[
    width=\columnwidth,
    colback=gray!5!white, 
    colframe=black, 
    title=R1 Trace Explanation Prompt, 
    fonttitle=\bfseries\large, 
    sharp corners, 
    boxrule=0.8pt,
    left=8pt, right=8pt, top=8pt, bottom=8pt
]
\large
\textbf{\{Problem\}}

\vspace{4pt}
\textbf{\{R1 trace\}}

\vspace{4pt}
\textbf{\{R1 answer\}}

\vspace{6pt}
You have answered the question correctly. Please provide a detailed explanation of the reasoning behind your answer.

\vspace{4pt}
\textbf{Explanation:}
\end{tcolorbox}







\subsection{Trace Length Analysis}
\begin{figure}[H]
    \centering
    \includegraphics[width=\columnwidth]{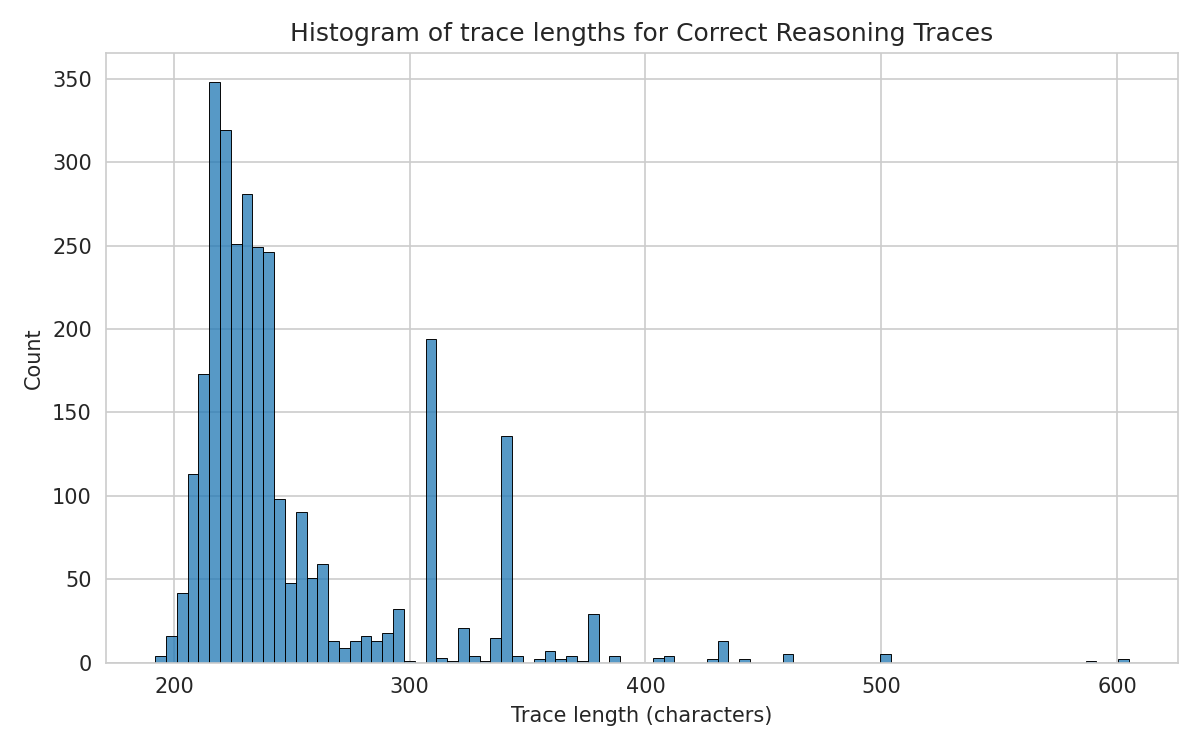}
    \caption{Distribution of trace lengths of algorithmically generated correct traces.}
    \label{fig:one}
\end{figure}

\begin{figure}[H]
    \centering
    \includegraphics[width=\columnwidth]{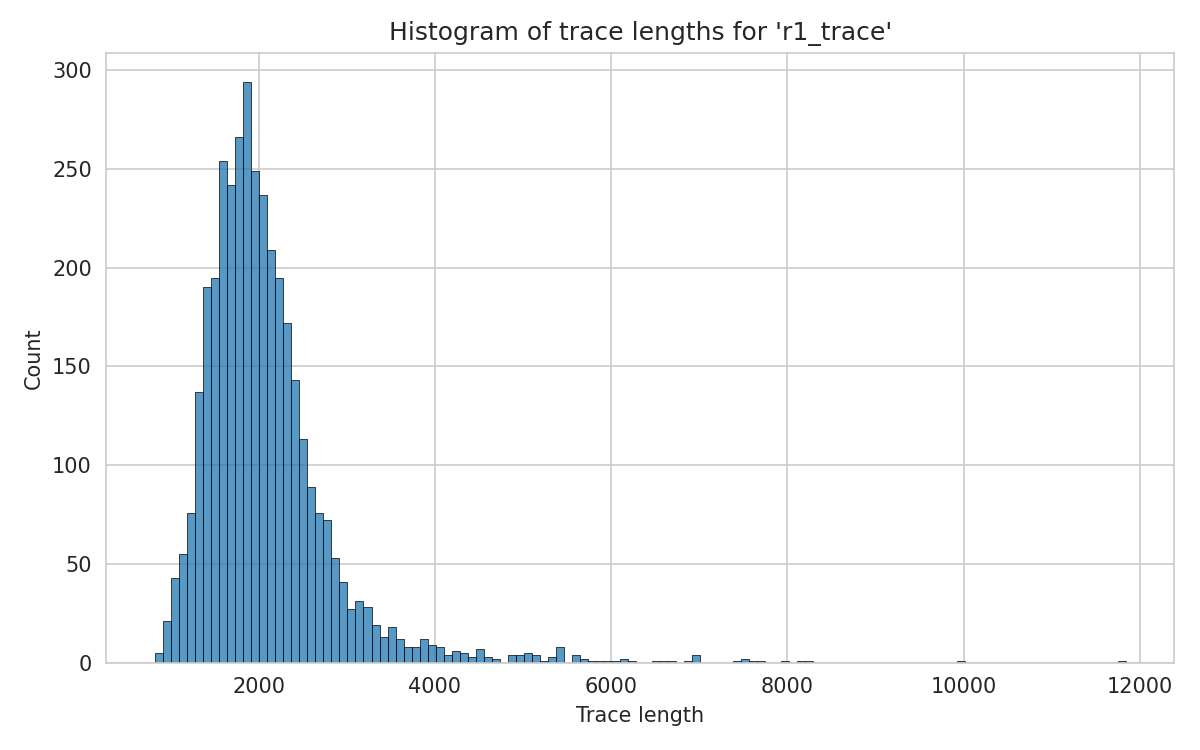}
    \caption{Distribution of trace lengths of R1 traces.}
    \label{fig:two}
\end{figure}

\begin{figure}[H]
    \centering
    \includegraphics[width=\columnwidth]{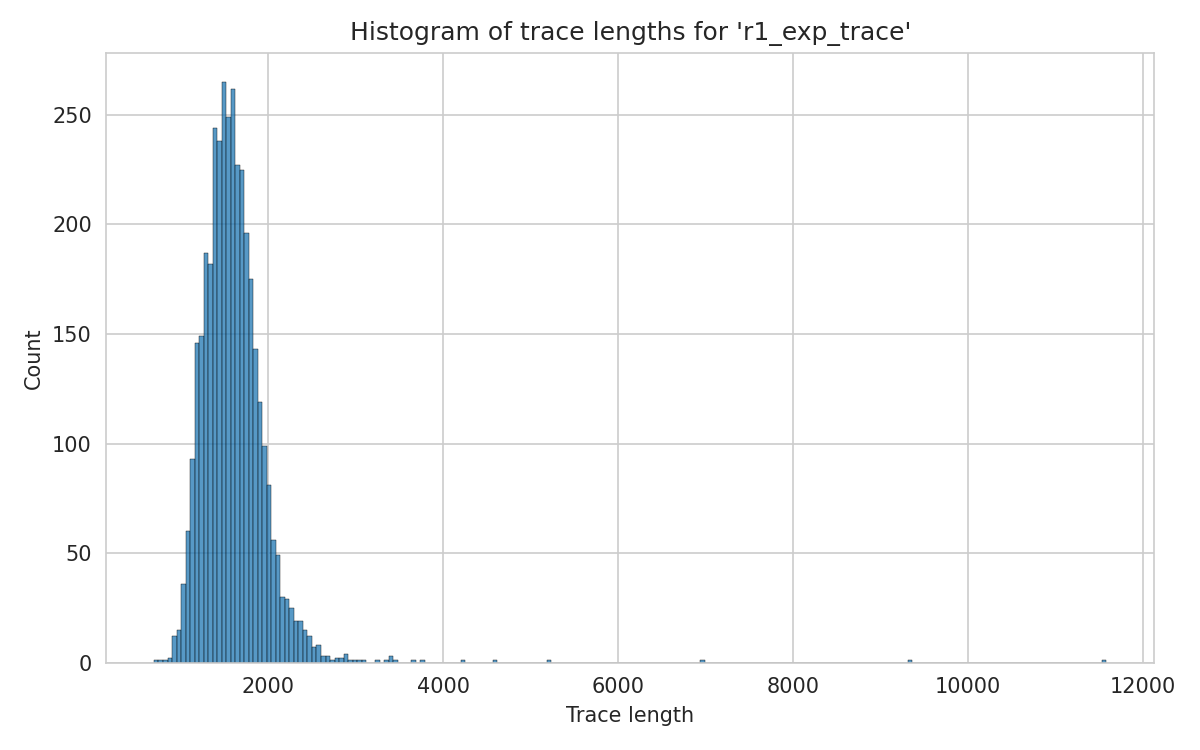}
    \caption{Distribution of trace lengths of post-hoc explanations of R1 traces.}
    \label{fig:three}
\end{figure}

\begin{figure}[H]
    \centering
    \includegraphics[width=\columnwidth]{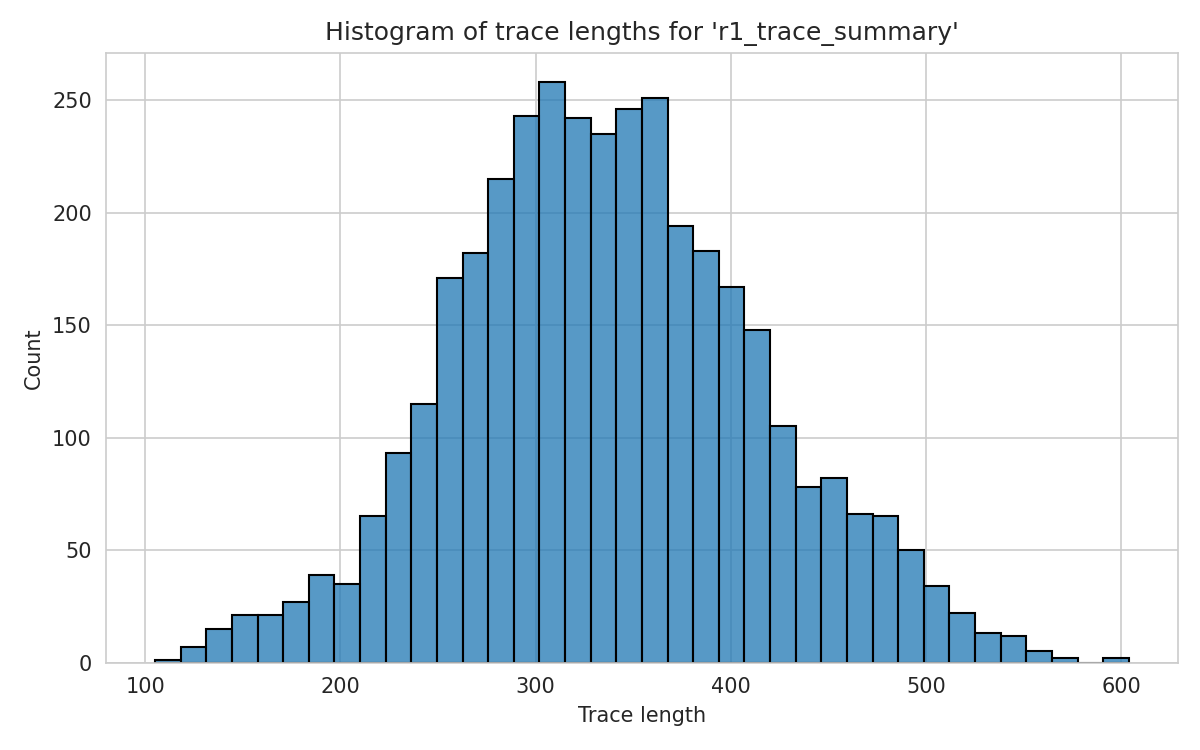}
    \caption{Distribution of trace lengths of summaries of R1 traces.}
    \label{fig:four}
\end{figure}

Figures \ref{fig:one}, \ref{fig:two}, \ref{fig:three} and \ref{fig:four} compare trace-length distributions for four reasoning trace types: algorithmically generated correct traces, R1 traces, post-hoc explanations of R1, and summarized R1 traces. It shows how R1 traces tend to be longer and more verbose, while summaries and explanations are more compact, affecting how much information users must process. 

\section{User Study}
\label{sec:appendix_user_study}
To evaluate the interpretability of reasoning traces generated by reasoning models, we conducted a set of structured user studies. Each participant was given a compensation of $\$12/hr$. The study was approved by the IRB protocol code 55564726. Each study followed the same sequence of steps, designed to ensure consistency across participants for each study. Below we outline the main components of the study design. 
\subsection{Human Participant Demographics}
We conducted four user studies with participants recruited through Prolific (all located in the United States). In general, the participant populations in all four studies were demographically similar, with no major differences in the age or education distribution, suggesting that the results in the studies are comparable and not driven by differences in the composition of the participants.
\paragraph{Education:}  
Participants spanned a range of educational backgrounds. Across all studies, the majority held an \emph{Undergraduate Degree} (roughly 45--55\% in each study), followed by \emph{Master's Degrees} (20--30\%), and a smaller proportion with \emph{PhDs or equivalent doctoral-level degrees} (10--15\%). A minority of participants reported \emph{High School}, \emph{Associate's Degree}, or \emph{Some College} as their highest level of education ($<$10\% each). These proportions were consistent across the four studies. 
\paragraph{Age:} The participants were distributed over a wide age range, with the largest groups being \emph{35--50 years old} (approximately 35--40\%) and \emph{51+ years old} (30--35\%). Younger age groups were represented to a lesser extent: \emph{26--34 years old} (20--25\%) and \emph{18--25 years old} (5--10\%). Again, these proportions were stable across studies.

\subsection{Consent and Statement}
\begin{figure}[ht]
    \centering
    \includegraphics[width=\columnwidth]{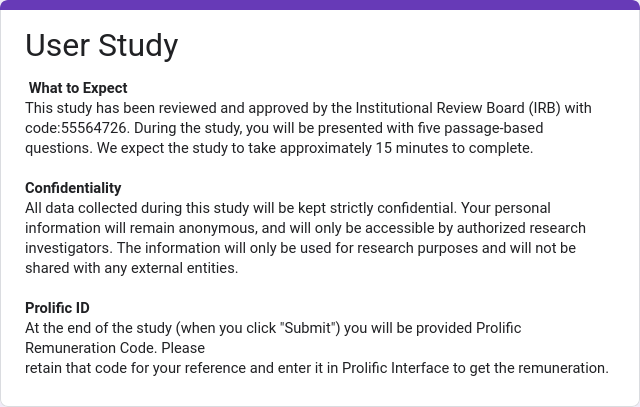}
    \caption{Consent statement shown to participants before starting the study.}
    \label{fig:consent_statement}
\end{figure}

Each participant began the study by reviewing and agreeing to a consent statement (Figure \ref{fig:consent_statement}). The statement explained the goals of the study, what participants would be asked to do, and how their data would be handled.

\subsection{Instructions}
\begin{figure}[t]
    \centering
    \includegraphics[width=\columnwidth]{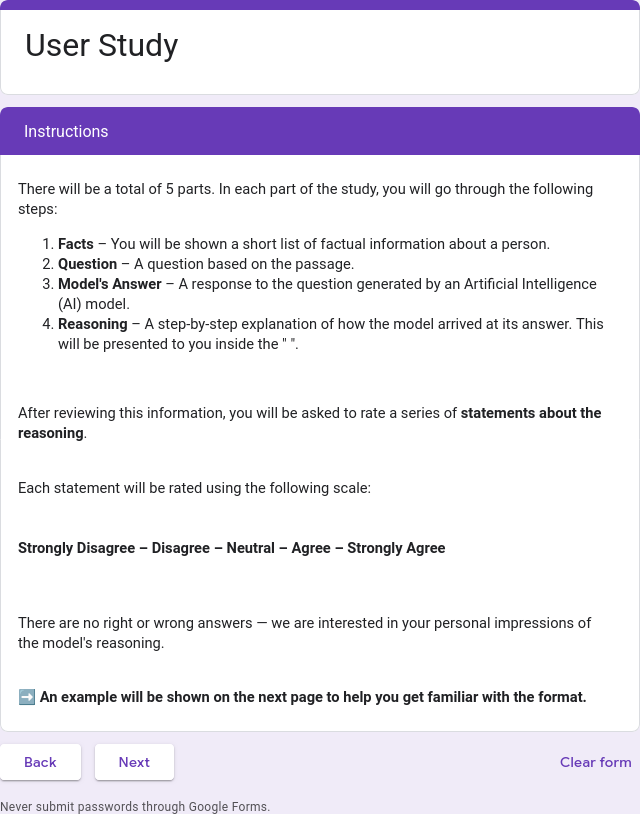}
    \caption{Instructions shown to participants before starting the study.}
        \label{fig:user_instructions}
\end{figure}

Participants were provided with detailed instructions describing the study structure (Figure \ref{fig:user_instructions}). Each of the five parts of the study followed the same format:
\begin{enumerate}
    \item \textbf{Facts:} A short list of factual statements about a person.  
    \item \textbf{Question:} A query based on the passage.  
    \item \textbf{Model's Answer:} The response generated by the AI model.  
    \item \textbf{Reasoning:} A step-by-step explanation of how the model arrived at its answer.  
\end{enumerate}

After reviewing this information, participants rated statements about the reasoning on a 5-point Likert scale (Strongly Disagree–Strongly Agree).  


\subsection{Q/A Task}
\begin{figure}[H]
    \centering
    \includegraphics[width=\columnwidth]{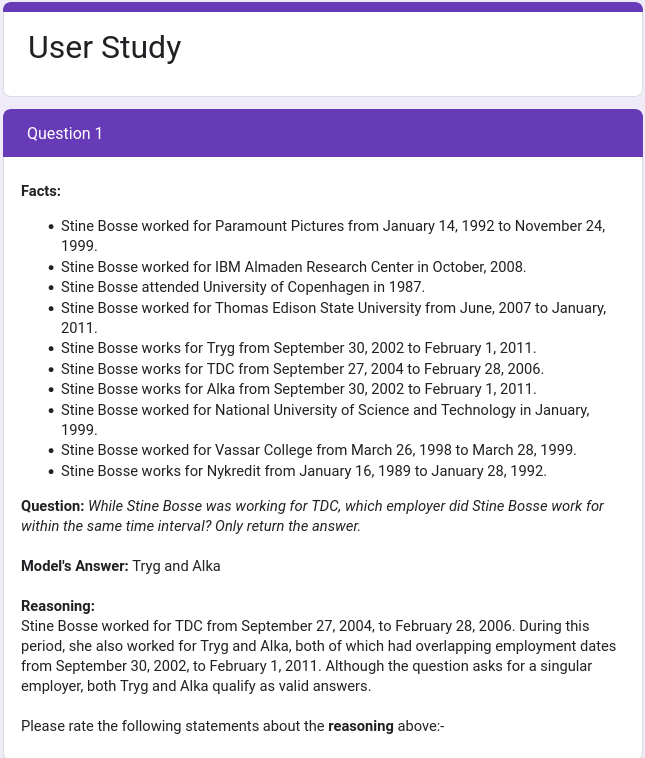}
        \caption{Task shown to participants.}
        \label{fig:user_task}
\end{figure}

\begin{figure}[H]
    \centering
    \includegraphics[width=\columnwidth]{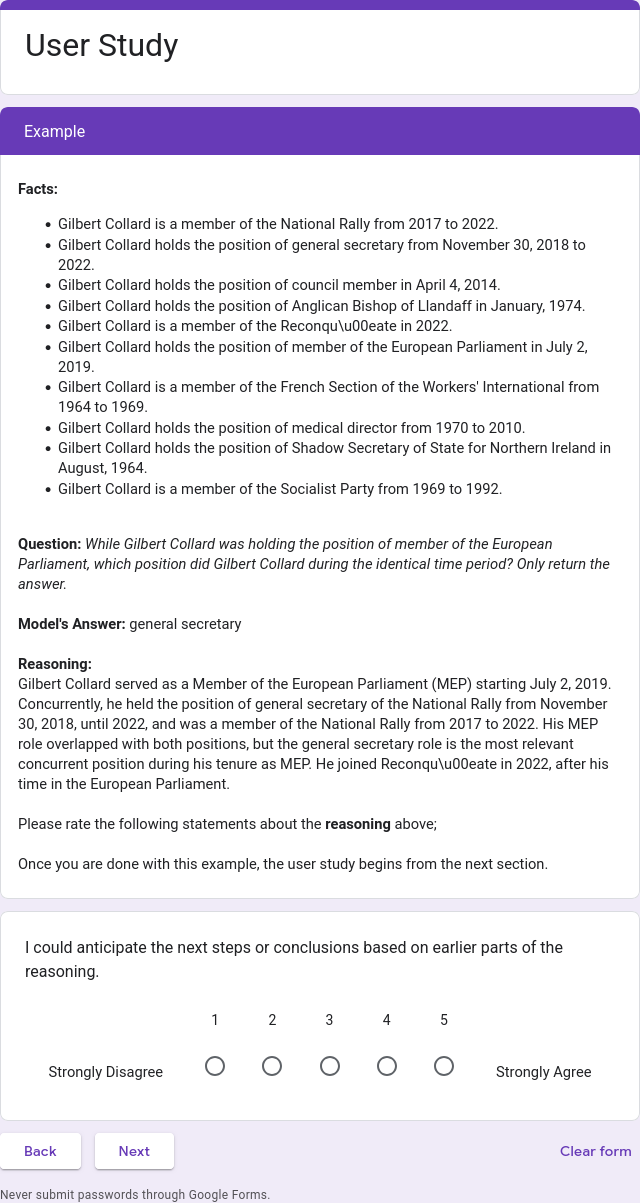}
        \caption{Example shown to participants.}
        \label{fig:user_example}
\end{figure}

Before beginning the main task, participants reviewed an example question and answer with reasoning (Figure \ref{fig:user_example}). Participants then completed five Q/A tasks of the same form as the example. Each task included a passage, model answer, reasoning trace, and associated questionnaire (Figure \ref{fig:user_task}).  

\subsection{Examples of Traces shown to Users}
\label{sec:app_user_traces}
Table.~\ref{table:trace-1}, \ref{table:trace-2}, \ref{table:trace-3}, \ref{table:trace-4}, \ref{table:trace-5} compares four representations of reasoning traces used in the study: the full R1 traces, summarized traces, post-hoc generated explanations, and the correct traces.

\renewcommand{\arraystretch}{1.3} 

\begin{table*}[t]
\centering
\caption{Comparison of R1 traces, summarized traces, post-hoc explanations, and correct traces for interpretability analysis: Instance 1}
\label{table:trace-1}
\small
\begin{tabular}{@{}p{0.18\textwidth}p{0.78\textwidth}@{}}
\toprule
\textbf{Type of Trace} & \textbf{Trace generated by model after SFT} \\
\midrule

DeepSeek R1 &
"Hmm, the user is asking about Gilbert Collard's positions during the time he was a member of the European Parliament. Let me carefully analyze the context provided. Looking at the timeline, Gilbert Collard became a member of the European Parliament on July 2, 2019. I need to find which other positions he held simultaneously during this period. From the context: He was general secretary of National Rally from November 30, 2018 to 2022. He was a member of National Rally from 2017 to 2022. He joined Reconquête in 2022. Since his European Parliament term started in July 2019 and overlapped with both his National Rally membership (2017-2022) and his general secretary position (2018-2022), both positions were held concurrently with his MEP role. However, the question specifically asks for positions held during the identical time period as his MEP tenure. Both 'member of National Rally' and 'general secretary' qualify, but I need to see which one is more precise. The general secretary role started before his MEP term (Nov 2018 vs July 2019) and ended at the same time (2022). The party membership also covers the entire period. But since the question mentions 'position,' both are valid, though general secretary is a specific appointed role. I notice the user's question says 'which position' (singular), so I should provide the most relevant concurrent position. Given that general secretary is an active leadership role within the party he belonged to during his MEP term, it seems the most directly relevant answer. The other positions either predate his MEP term (like his medical director role ending in 2010) or are from different periods (like his Socialist Party membership ending in 1992). The Anglican Bishop and Shadow Secretary positions seem like possible data errors as they don't align with his known political career. Therefore, while holding the MEP position from July 2019 onward, Gilbert Collard was simultaneously the general secretary of National Rally until 2022." \\

\hline
LLM-generated R1 Summary &
"Gilbert Collard served as a Member of the European Parliament (MEP) starting July 2, 2019. Concurrently, he held the position of general secretary of the National Rally from November 30, 2018, until 2022, and was a member of the National Rally from 2017 to 2022. His MEP role overlapped with both positions, but the general secretary role is the most relevant concurrent position during his tenure as MEP. He joined Reconquête in 2022, after his time in the European Parliament." \\

\hline
LLM-generated R1 post-hoc Explanation &
"Alright, let me break down how I arrived at the answer step by step. The user wants to know which position Gilbert Collard held simultaneously while serving as a Member of the European Parliament (MEP). First, I identify the key timeframe: Collard became an MEP on July 2, 2019. This is my anchor point. Now I need to examine all his other positions to see which overlapped exactly with this period. Looking through the data: His role as general secretary ran from November 30, 2018 to 2022. Since July 2019 falls between these dates, this overlaps. His National Rally membership (2017-2022) also covers the MEP period, but the question specifically asks for 'position' - implying a formal role, not just membership. Other positions either ended before 2019 (like medical director ending in 2010) or started after he left the MEP role (Reconquête membership in 2022). Notably, some entries seem implausible based on known facts (like Anglican Bishop in 1974 or Shadow Secretary in 1964), so I disregard those as likely errors. Therefore, the only verifiable formal position held concurrently with his MEP term is general secretary of National Rally. The timeline alignment is clear: he held both roles continuously from July 2019 until 2022." \\

\hline
Verifiably Correct Trace &
"The temporal relation between the event in question and the event in context is: during. I need to use the following facts to answer the question: \{'Gilbert Collard holds the position of general secretary from November 30, 2018 to 2022.'\}" \\

\hline
Verifiably Incorrect Trace &
\textcolor{black}{\small{"The temporal relation between the event in question and the event in context is: equal. I need to use the following facts to answer the question: Gilbert Collard is a member of the Socialist Party from 1969 to 1992."}} \\

\bottomrule
\end{tabular}
\end{table*}

\begin{table*}[t]
\centering
\caption{\textcolor{black}{Comparison of R1 traces, summarized traces, post-hoc explanations, and correct traces for interpretability analysis: Instance 2}}
\label{table:trace-2}
\small
\begin{tabular}{@{}p{0.18\textwidth}p{0.78\textwidth}@{}}
\toprule
\textbf{Type of Trace} & \textbf{Trace generated by model after SFT} \\
\midrule

DeepSeek R1 &
\textcolor{black}{\small{"Okay, I'm trying to figure out this question about Stine Bosse's work history. The question is asking which employer she worked for at the same time she was working for TDC. Let me look at the timeline carefully. First, I see that Stine worked for TDC from September 27, 2004 to February 28, 2006. So that's the time period we're focusing on. Now I need to check which other employers she worked for during this exact timeframe. Looking at the list: Paramount Pictures was from 1992 to 1999, which is before TDC. IBM Almaden was in October 2008, after TDC. University of Copenhagen was in 1987, way before. Thomas Edison State University was from 2007-2011, after TDC. Tryg was from 2002-2011, which overlaps with TDC's timeframe. Alka was also from 2002-2011, overlapping too. National University of Science and Technology was in January 1999, before. Vassar College was from 1998-1999, also before. Nykredit was from 1989-1992, way before. So both Tryg and Alka have employment periods that completely cover the TDC timeframe. Since the question asks which employer she worked for "within the same time interval" as TDC, both Tryg and Alka qualify. But I notice something interesting - both Tryg and Alka have identical employment dates: September 30, 2002 to February 1, 2011. This exactly encompasses the TDC period. The question says "which employer" (singular) but technically there are two overlapping employers. However, since both are valid answers and the context doesn't specify choosing only one, I should list both. But the instruction says "only return the answer" so I'll provide both names since they both meet the criteria."}} \\

\hline
LLM-generated R1 Summary &
\textcolor{black}{\small{"Stine Bosse worked for TDC from September 27, 2004, to February 28, 2006. During this period, she also worked for Tryg and Alka, both of which had overlapping employment dates from September 30, 2002, to February 1, 2011. Although the question asks for a singular employer, both Tryg and Alka qualify as valid answers."}} \\

\hline
LLM-generated R1 post-hoc Explanation &
\textcolor{black}{\small{"Alright, let me break this down. The user wants a clear explanation of how I determined which employers Stine Bosse worked for simultaneously with TDC. First, I recall that the key task was cross-referencing employment periods. The critical dates are TDC's timeframe: September 27, 2004, to February 28, 2006. I scanned all other employers: Paramount Pictures (1992-1999) ended before TDC started. IBM (2008) and Thomas Edison (2007-2011) started after TDC ended. University of Copenhagen (1987) and Nykredit (1989-1992) were too early. National University (Jan 1999) and Vassar College (1998-1999) didn't overlap. But Tryg and Alka both spanned 2002-2011—fully encompassing TDC's duration. The user seems satisfied with the initial answer but wants deeper reasoning. They might be testing my ability to logically parse timelines or preparing for similar data analysis tasks. What's interesting is how Tryg/Alka share identical dates. This suggests either parallel employment or data entry quirks, but the question only cares about temporal overlap—which both satisfy. I'll emphasize the date comparisons clearly, avoiding jargon. Since the user appreciated conciseness before, I'll keep the explanation tight but thorough."}} \\

\hline
Verifiably Correct Trace &
\textcolor{black}{\small{"The temporal relation between the event in question and the event in context is: during. I need to use the following facts to answer the question: ['Stine Bosse works for Tryg from September 30, 2002 to February 1, 2011.', 'Stine Bosse works for Alka from September 30, 2002 to February 1, 2011.']."}} \\

\hline
Verifiably Incorrect Trace &
\textcolor{black}{\small{"The temporal relation between the event in question and the event in context is: mix. I need to use the following facts to answer the question: Stine Bosse worked for Thomas Edison State University from June, 2007 to January, 2011."}} \\

\bottomrule
\end{tabular}
\end{table*}

\begin{table*}[t]
\centering
\caption{\textcolor{black}{Comparison of R1 traces, summarized traces, post-hoc explanations, and correct traces for interpretability analysis: Instance 3}}
\label{table:trace-3}
\small
\begin{tabular}{@{}p{0.18\textwidth}p{0.78\textwidth}@{}}
\toprule
\textbf{Type of Trace} & \textbf{Trace generated by model after SFT} \\
\midrule

DeepSeek R1 &
\textcolor{black}{\small{"Alright, so I need to figure out which position Thomas Wenski held at the same time he was the auxiliary bishop. Let me start by looking at the context provided. First, I see that Thomas Wenski became the auxiliary bishop on June 24, 1997. That's a specific date, so I should look for other positions he held around that same time. Scanning through the list, I notice several other positions with dates: Titular bishop on June 24, 1997. Catholic bishop on September 3, 1997. Minister for Children (Denmark) on June 2, 1998. Head of the Security Service of Ukraine on April 13, 1999. Director of the Federal Security Service in May 1999. President of Vietnam from 1999 to 2001. Deputy President of the Supreme Court of the UK in November 2002. Parliamentary Under-Secretary of State for Employment from November to December 2004. Diocesan bishop on November 13, 2004. Wait a minute, I see that both the auxiliary bishop and titular bishop positions started on the exact same date: June 24, 1997. That means that while he was holding the position of auxiliary bishop, he was also the titular bishop at the same time. The other positions have different dates, either before or after June 24, 1997, so they wouldn't overlap. For example, the Catholic bishop role started a few months later in September, and the Minister for Children role started almost a year later in June 1998. All the other positions are even later, so they don't coincide with his time as auxiliary bishop. Therefore, the position Thomas Wenski held during the same time period as auxiliary bishop is the titular bishop."}} \\

\hline
LLM-generated R1 Summary &
\textcolor{black}{\small{"Thomas Wenski became auxiliary bishop on June 24, 1997, simultaneously holding the position of titular bishop, which also started on the same date. Other roles he held, such as Catholic bishop and Minister for Children, began later, indicating no overlap with his auxiliary bishop tenure. Thus, the only concurrent position was titular bishop."}} \\

\hline
LLM-generated R1 post-hoc Explanation &
\textcolor{black}{\small{"Okay, the user wants me to explain my reasoning for the previous answer about Thomas Wenski holding positions simultaneously. I correctly identified that he was both auxiliary bishop and titular bishop on the same start date. Hmm, looking at the context again, every position entry has clear dates. The key was noticing that both "auxiliary bishop" and "titular bishop" shared the exact start date of June 24, 1997. That's the critical overlap. The user seems satisfied with the answer but wants a clear breakdown. They're probably verifying if I understood the temporal reasoning or just guessed. Maybe they're testing how I handle simultaneous appointments. I should emphasize the date matching in the explanation. Other positions either started later (like Catholic bishop in September) or much later (1998-2004), so they couldn't overlap with the June 1997 auxiliary bishop role. The explanation should stay concise but highlight why only titular bishop fits. No need to mention all other positions - just contrast with one nearby example (like September's Catholic bishop role) to show the time gap."}} \\

\hline
Verifiably Correct Trace &
\textcolor{black}{\small{"The temporal relation between the event in question and the event in context is: equal. I need to use the following facts to answer the question: ['Thomas Wenski holds the position of titular bishop in June 24, 1997.']"}} \\

\hline
Verifiably Incorrect Trace &
\textcolor{black}{\small{"The temporal relation between the event in question and the event in context is: overlap. I need to use the following facts to answer the question: Thomas Wenski holds the position of auxiliary bishop in June 24, 1997."}} \\

\bottomrule
\end{tabular}
\end{table*}

\begin{table*}[t]
\centering
\caption{\textcolor{black}{Comparison of R1 traces, summarized traces, post-hoc explanations, and correct traces for interpretability analysis: Instance 4}}
\label{table:trace-4}
\small
\begin{tabular}{@{}p{0.18\textwidth}p{0.78\textwidth}@{}}
\toprule
\textbf{Type of Trace} & \textbf{Trace generated by model after SFT} \\
\midrule

DeepSeek R1 &
\textcolor{black}{\small{"Alright, so I need to figure out which position Kalervo Kummola held while he was the vice president. Let me break this down step by step to make sure I get it right. First, I'll look at the timeline for when he was vice president. The context says that Kalervo Kummola was the vice president from 2003 to 2021. That's a pretty long period—almost 18 years. So, any position he held during this time frame would be the answer. Now, I'll go through each of the positions listed in the context and check their dates to see which ones overlap with his vice presidency period. President of Romania in 2002: This is before 2003, so it doesn't overlap with his vice presidency. Foreign minister in November 1998: Also before 2003, so no overlap. United States Deputy Secretary of Labor in 1999: Before 2003, so no. Minister for Nordic Co-operation in July 1998: Again, before 2003. Anglican Bishop of Salisbury in 1999: Still before 2003. Confederate States Secretary of State in September 20, 2001: This is in 2001, which is before 2003. Lord Lieutenant of Ireland in August 1999: Before 2003. Board member from 1998 to 2021: This position starts in 1998 and ends in 2021. Since his vice presidency is from 2003 to 2021, there's an overlap from 2003 to 2021. Chairperson from 1997 to January 7, 2016: This ends in January 2016, which is within the vice presidency period (2003-2021). So, he was chairperson until 2016 while being vice president. Both positions overlap with his vice presidency period. However, the question asks which position he held within the same time interval as being vice president. Both board member and chairperson fit this criterion."}} \\

\hline
LLM-generated R1 Summary &
\textcolor{black}{\small{"Kalervo Kummola served as vice president from 2003 to 2021. During this period, he also held the positions of board member (1998-2021) and chairperson (1997-January 2016). Both roles overlapped with his vice presidency, making them relevant to the inquiry. Thus, the answer includes both positions."}} \\

\hline
LLM-generated R1 post-hoc Explanation &
\textcolor{black}{\small{"Okay, the user wants me to explain my reasoning behind identifying which positions Kalervo Kummola held while serving as vice president. Let me recall the context clearly: the vice presidency spanned from 2003 to 2021, and I need to find overlapping roles. First, I'll mentally scan all listed positions. Most roles like President of Romania (2002) or foreign minister (1998) clearly fall outside the 2003-2021 window. The Confederate States Secretary role in 2001 is too early too. Then two positions catch my eye: board member (1998-2021) and chairperson (1997-Jan 2016). Both end dates extend into the vice presidency period. For board member, the entire 2003-2021 duration is covered under its 1998-2021 tenure. For chairperson, even though it started earlier, it actively overlapped from 2003 until January 2016—which is well within the vice presidency timeline. The user seems satisfied with the initial answer but wants deeper clarity. Since all other positions are time-bound snapshots that don't touch 2003-2021, only these two long-term roles qualify."}} \\

\hline
Verifiably Correct Trace &
\textcolor{black}{\small{"The temporal relation between the event in question and the event in context is: mix. I need to use the following facts to answer the question: ['Kalervo Kummola holds the position of board member from 1998 to 2021.', 'Kalervo Kummola holds the position of chairperson from 1997 to January 7, 2016.']"}} \\

\hline
Verifiably Incorrect Trace &
\textcolor{black}{\small{"The temporal relation between the event in question and the event in context is: equal. I need to use the following facts to answer the question: Kalervo Kummola holds the position of President of Romania in 2002."}} \\

\bottomrule
\end{tabular}
\end{table*}

\begin{table*}[t]
\centering
\caption{\textcolor{black}{Comparison of R1 traces, summarized traces, post-hoc explanations, and correct traces for interpretability analysis: Instance 5}}
\label{table:trace-5}
\small
\begin{tabular}{@{}p{0.18\textwidth}p{0.78\textwidth}@{}}
\toprule
\textbf{Type of Trace} & \textbf{Trace generated by model after SFT} \\
\midrule

DeepSeek R1 &
\textcolor{black}{\small{"Alright, let's tackle this problem step by step. The question is asking which position Pierre-Joseph-Olivier Chauveau held simultaneously while he was a member of the House of Commons of Canada. So, first, I need to figure out the timeframe during which he held that position. Looking at the context provided, Pierre-Joseph-Olivier Chauveau was a member of the House of Commons of Canada from September 20, 1867, to January 22, 1874. That's a period of about six and a half years. Now, I need to see which other positions he held during this same timeframe. Speaker of the New Zealand House of Representatives: July 1874 — after his term ended, no overlap. Speaker of the Senate: February 21, 1873, to January 8, 1874 — overlaps from Feb 1873 to Jan 1874. Catholic bishop: March 1867 — before his term started, no overlap. President of Singapore: May 9, 1874 — after his term, no overlap. Prime Minister of Albania: August 1874 — after his term, no overlap. Prime Minister of New Zealand: October 11, 1874 — after his term, no overlap. Minister of Works: February 1874 — after January 22, 1874, no overlap. Secretary of Education: June 9, 1867 — before September 1867, no overlap. Premier of Quebec: July 15, 1867, to February 25, 1873 — overlaps from July 1867 to Feb 1873. Therefore, the positions he held simultaneously while being a member of the House of Commons of Canada are Premier of Quebec and Speaker of the Senate."}} \\

\hline
LLM-generated R1 Summary &
\textcolor{black}{\small{"Pierre-Joseph-Olivier Chauveau served as a member of the House of Commons of Canada from September 20, 1867, to January 22, 1874. During this period, he simultaneously held the positions of Premier of Quebec (July 15, 1867 - February 25, 1873) and Speaker of the Senate (February 21, 1873 - January 8, 1874). Thus, his overlapping roles were as Premier of Quebec and Speaker of the Senate while serving in the House of Commons."}} \\

\hline
LLM-generated R1 post-hoc Explanation &
\textcolor{black}{\small{"Alright, the user wants a detailed explanation of how I arrived at the correct answer. First, I establish the timeframe for his House of Commons membership: September 20, 1867, to January 22, 1874. Then I check all other positions for overlap. Speaker of the Senate (Feb 21, 1873 - Jan 8, 1874) overlaps briefly. Premier of Quebec (July 15, 1867 - February 25, 1873) overlaps for over five years. All other positions fall either before September 1867 or after January 1874. The key nuance is that in Canada, the House of Commons and Senate are separate chambers — a person cannot simultaneously be a member of both. So while the dates suggest two overlaps, the constitutional constraint makes only Premier of Quebec a valid concurrent role, since a provincial premiership and federal MP role can coexist. That explains why Premier of Quebec is the correct simultaneous position."}} \\

\hline
Verifiably Correct Trace &
\textcolor{black}{\small{"The temporal relation between the event in question and the event in context is: mix. I need to use the following facts to answer the question: ['Pierre-Joseph-Olivier Chauveau holds the position of Speaker of the Senate from February 21, 1873 to January 8, 1874.', 'Pierre-Joseph-Olivier Chauveau holds the position of Premier of Quebec from July 15, 1867 to February 25, 1873.']"}} \\

\hline
Verifiably Incorrect Trace &
\textcolor{black}{\small{"The temporal relation between the event in question and the event in context is: during. I need to use the following facts to answer the question: Pierre-Joseph-Olivier Chauveau holds the position of Prime Minister of Albania in August, 1874."}} \\

\bottomrule
\end{tabular}
\end{table*}

\end{document}